\documentclass[11pt]{article}
\usepackage[preprint]{acl}
\usepackage{times}
\usepackage{latexsym}
\usepackage[T1]{fontenc}
\usepackage[utf8]{inputenc}
\usepackage{microtype}
\usepackage{inconsolata}
\usepackage{amsmath,amssymb}
\usepackage{graphicx}
\usepackage{placeins}
\graphicspath{{./}}
\usepackage{booktabs}
\usepackage{longtable}
\usepackage{newunicodechar}
\newunicodechar{’}{'}
\newunicodechar{–}{--}
\title{Think or Remember: The Exchange Rate of Agentic Inference}
\author{Juan Wisznia\textsuperscript{1} \qquad Santiago Luis Groba Alonso\textsuperscript{2}\\
  \textbf{Mat\'ias G\'omez Taddeo\textsuperscript{2} \qquad Gonzalo Zabaleta\textsuperscript{2} \qquad Joaqu\'in Cullen\textsuperscript{2} \qquad Luciano Del Corro\textsuperscript{1}}\\
  \textsuperscript{1}ELIAS Lab, Escuela de Ingenier\'ia, Universidad de San Andr\'es, Buenos Aires, Argentina\\
  \textsuperscript{2}Escuela de Ingenier\'ia, Universidad de San Andr\'es, Buenos Aires, Argentina\\
  \texttt{\{jwisznia,grobaalonsos,mgomeztaddeo,gzabaletaferrando,cullenj,delcorrol\}@udesa.edu.ar}}
\begin{document}
\maketitle
\begin{abstract}
Agents built on reasoning models resend their growing transcript at every call. Two transacting parties, not the model, decide what that costs: the client keeps or edits messages; the server decides whether earlier reasoning returns in the prompt and what its cache reuses. We define a cost model whose counts match an instrumented engine exactly; the architecture charges $c$ per token processed and $\beta$ per pair attended. Their ratio $P^*=c/\beta$ is an exchange rate between thinking and remembering: per token, processing dominates below this context length, a thinking regime, and attention above it, a memory regime. A policy change has a mechanical cost, incurred even if the agent writes the same tokens, and a behavioral one: it writes different tokens, changing cost and outcome. Serving alone can flip which policy is cheaper: on recorded Qwen3-8B ScienceWorld runs, dropping earlier reasoning costs 1.74 times keeping it when the server reuses generated states and 0.34 times when it reuses nothing. In live reruns, dropping it cuts correct HotpotQA answers by 9--23 percentage points on four models, at lower point-estimate cost. Comparing context policies means comparing systems: architecture and serving set the cost of the history kept and the work it induces.
\end{abstract}
\section{Introduction}
\label{sec:econ-intro}

Reasoning models write a trace of tokens before every answer, which their providers call thinking, and agents built on them run for hundreds of calls, sending the whole transcript back each time and paying to process whatever the prefix cache \citep{kwon23vllm} does not already hold. Two parties decide what that transcript costs. The client keeps, summarizes or edits the visible history. The server decides whether the reasoning of earlier turns returns in the prompt, which we call Keep when it does and Drop when it does not, and which of its computations the cache reuses; what the model is shown and what the server recomputes are separate choices. Whether to keep or drop the reasoning, or whether to compact the history, is not a property of the model but a decision made by two transacting parties.

Recent methods report what client policies save in tokens, dollars or FLOPs \citep{hoy2026continuouscontextmanagement,satish2026tokensavingssystematicstudy,shao2026contextlanguagemodels}, usually under one serving convention (Section~\ref{sec:econ-related}). However, a token's cost depends on where in the context it sits and on whether the server already holds the work done on it, so token counts alone do not price a policy. In this work we formalize a run's total cost as a function of the model's architecture and of the choices of both transacting parties.

Every token an agent carries forward is a choice between two ways of having the same information. A token held in context is paid for by attention, a little by every later token for as long as the run lasts. A token dropped is paid for when it is needed again: by producing it once more, and by rebuilding whatever followed it in the cache. At equal base prices for input and generation, a run costs $C=cN+\beta Q$, where $N$ counts the positions processed, repeats included, $Q$ the pairs attended, and the two prices $c$ and $\beta$ are fixed by the architecture (Section~\ref{sec:econ-prices}). Their ratio $P^*=c/\beta$ is the exchange rate between thinking and remembering: the context length past which a token pays more to look back than to pass through the weights, 124K tokens on Qwen3.6-27B, where 16 of 64 layers attend, and 30K on a synthetic all-attention variant of it. Tokens evaluated with fewer than $P^*$ tokens of context are in a thinking regime; beyond $P^*$, they are in a memory regime, as when a task needs a long context to be answered correctly. A whole run compares $Q/N$, the mean keys per evaluated position, with $P^*$. Serving sets the rate for thinking again: a cache rule $\rho$ decides whether generated states are reused or reprocessed when they return, and a decoding premium $w$ prices each generated token at $wc$ (Equation~\ref{eq:econ-regeneration-rate}).

With generated-state reuse, a reasoning trace of $r$ tokens that follows $H$ tokens of history and precedes an answer of $a$ tokens charges every later evaluation $\beta r$ for attending to it, while dropping it rebuilds the answer once; the two are equal after $L^*$ later evaluated positions (Equation~\ref{eq:econ-horizon}). On Qwen3.6-27B at $H=20$K, a 2,000-token trace that ends in a 7-token answer is worth keeping for about 500 more evaluated tokens; the same trace ending in a 4,000-token command is worth keeping for about 290K. Below the horizon, retaining the block is mechanically cheaper; above it, deleting it is, given an exact reusable prefix and no later edit or eviction.

These prices hold the agent's work fixed, but the work changes with the policy. Shown a different history, the model writes a different continuation, restating a plan it can no longer see, repeating one it can, or stopping. We therefore split every change of policy in two (Section~\ref{sec:econ-split}). The mechanical part is what the change costs if the agent writes exactly the same tokens: tokens read again to rebuild the cache, and later tokens attending to a different context. Replaying the recorded run with the other policy's prompts computes it exactly. The behavioral part is what the change costs because the agent writes different tokens, which only rerunning the agent under the other policy from the same point reveals. An intervention pays for itself when its behavioral saving exceeds its mechanical expense.

The model's token, pair and cache counts match an instrumented vLLM exactly on all 648 requests of a dense and a hybrid model. Prices alone can reverse a comparison: on the same recorded Qwen3-8B ScienceWorld runs, with the agent's tokens unchanged, dropping earlier reasoning costs 1.74 times keeping it when the server reuses the states of generated tokens and 0.34 times when it reuses nothing (Section~\ref{sec:econ-repricing}). The timing of a deletion matters more than its size: on recorded hosted runs with the same reuse, a sliding window that keeps the latest five reasoning blocks costs about four times keeping everything, because each advance of the window rebuilds the cache behind it, while deleting each block at once stays within 2\% of keeping (Section~\ref{sec:econ-mechanical-policies}).

Behavior then moves both the cost and the outcome. Rerun live from the same points, dropping earlier reasoning loses 9 to 23 percentage points of fully correct answers on title-hidden HotpotQA across four models at lower point-estimate cost (intervals include equal cost), yet lifts gpt-oss-20b on ScienceWorld from 3 to 20 solved tasks of 88 (Section~\ref{sec:econ-live}). Carrying history across chained ScienceWorld tasks under Keep costs 0.60 of fresh sessions but solves 17 of 96 new tasks against 30 (Section~\ref{sec:econ-sequential}). Whether model-directed editing or automatic summarization is cheaper depends on the server's retention rule and the cohort (Section~\ref{sec:econ-repayment}).

The per-token accounting follows \citet{kaplan2020scaling} and CLM \citep{shao2026contextlanguagemodels}. We add three things: an account in which serving enters alongside architecture and moves the exchange rate (Section~\ref{sec:econ-theory}); the exact split of a policy change into mechanical and behavioral parts, with the condition under which an intervention repays (Section~\ref{sec:econ-split}); and evidence from six open models in fork studies, two more in the engine validation and one hosted model, on ScienceWorld, HotpotQA and chained tasks (Section~\ref{sec:econ-experiments}). An otherwise controlled comparison of client policies identifies their effect at a shared server setting even if it is unreported. Reporting that setting lets readers check this condition and assess transfer: specify the architecture, cache rule, earlier reasoning shown and who set it, and each call's prompt, cached and completion counts.
\section{The cost of a run}
\label{sec:econ-theory}

\begin{figure*}[t]
\centering
\includegraphics[width=.95\textwidth]{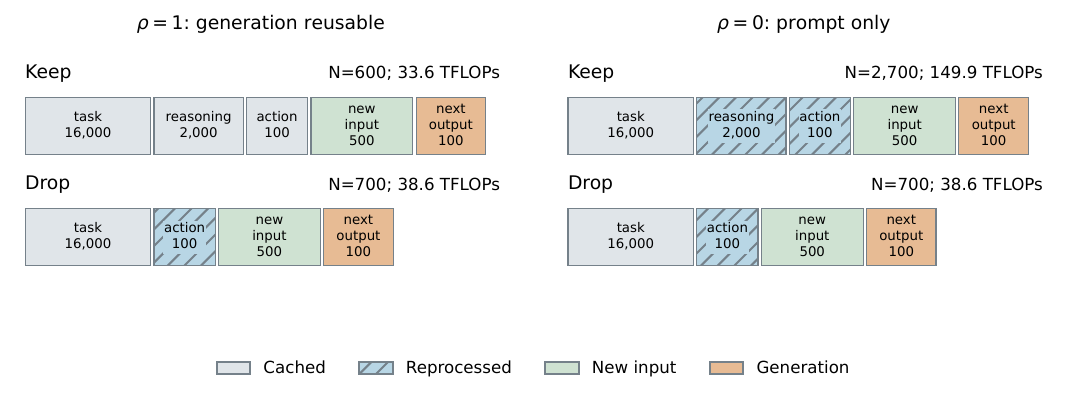}
\caption{\textbf{Serving changes the price of the same history.} A call adds 2,000 reasoning and 100 action tokens to a cached 16,000-token task; the next call adds 500 input and 100 generated tokens. Hatching marks rebuilding: Drop rebuilds the action under $\rho=1$, and Keep also rebuilds reasoning under $\rho=0$. Qwen3.6-27B prices, $w=1$, token-granular reuse; widths are schematic.}
\label{fig:econ-cache-blocks}
\end{figure*}

\subsection{Calls, tokens, and two prices}
\label{sec:econ-prices}

At call $t$, the server renders a prompt of $P_t$ tokens. The first $R_t$ positions are reusable from its cache, so the engine processes (evaluates) the remaining $P_t-R_t$ input positions, then $G_t$ positions during generation. A position processed again counts again; a cached position is not processed but remains a key for later positions. We call the generated text and external inputs from a chosen call onward its continuation, $\xi$. A context policy $\pi$ specifies which parts of this history return in each prompt, including the server's treatment of earlier reasoning: Keep returns it, Drop removes it.

Each evaluated position incurs a base cost $c$ for processing through the model's projection layers, plus an attention cost $\beta$ for each key it attends to. Architecture determines these two prices.

Input and generation use the same projections but different hardware paths: prefill shares weight reads across many input tokens, while decoding produces one token per sequence per step and can be bandwidth-limited at small batches. A decoding premium $w$ prices a generation position's projection work at $wc$, with attention still priced at $\beta$. We examine $w=1$ through $w=5$; Appendix~\ref{app:econ-timing} reports how measured decoding time depends on batch size and context length.

The cache rule $\rho\in\{0,1\}$ says what the server keeps of what the model wrote: under generated-state reuse ($\rho=1$) the states of generated tokens are reusable on later calls; under prompt-only reuse ($\rho=0$) they are reprocessed when they return as input. A system with no reuse at all has $R_t=0$ at every call; we treat it as a third case. Let $N_\rho$ count all processed positions, $Q_\rho$ all attended pairs and $G$ the generation positions. Since $cN_\rho$ already charges generation at the base rate, the additional premium is $(w-1)cG$. The run costs

\begin{equation}
 \boxed{C_{\rho,w}(\pi;\xi)=cN_\rho+\beta Q_\rho+(w-1)cG.}
 \label{eq:econ-cost}
\end{equation}

Under contiguous prefix reuse (the actual $R_t$ also depends on token identity, block alignment and eviction), the counts are

\begin{align}
 N_\rho&=\sum_t(P_t-R_t+G_t),\label{eq:econ-count-n}\\
 Q_\rho&=\sum_t\left[\sum_{i=R_t+1}^{P_t}i+\sum_{j=1}^{G_t}(P_t+j)\right],\label{eq:econ-count-q}
\end{align}

with $G=\sum_tG_t$. At evaluated position $i$, causal attention covers $i$ keys, giving the inner sums. The GPU validation counts $G_t=O_t-1$ of $O_t$ emitted tokens, as the first comes from prefill; the fork-study prices use $O_t$ (Appendix~\ref{app:econ-derivations}).

Text retention and computation reuse are distinct. In Figure~\ref{fig:econ-cache-blocks}, removing earlier reasoning shortens the next prompt but forces its surviving answer to be processed again (Appendix~\ref{app:econ-worked-cache}).

At $w=1$, Equation~\ref{eq:econ-cost} prices token-processing and full-attention arithmetic; the decomposition follows \citet[Table~1 and Eq.~2.2]{kaplan2020scaling}. Appendix~\ref{app:econ-coefficients} derives the architectural coefficients, and Appendix~\ref{app:econ-accounting} gives the additional output-head and recurrent-state counts used in the GPU validation.

\par\vspace{0pt plus 6\baselineskip}\penalty-200\vspace{0pt plus -6\baselineskip}
\subsection{Exchange rates and the cost of carrying history}

The ratio of the two prices, in attended pairs per token processed, is

\begin{equation}
 P^*=c/\beta.
 \label{eq:econ-pstar}
\end{equation}

\begin{figure}[t]
\centering
\includegraphics[width=\columnwidth]{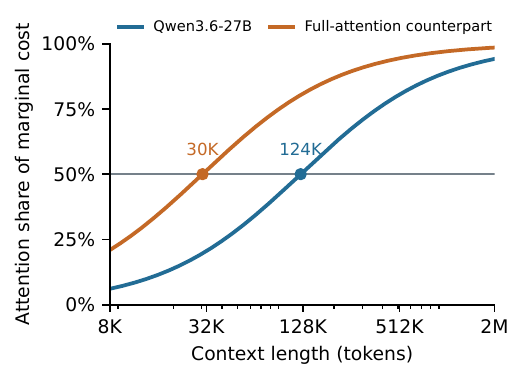}
\caption{\textbf{The architectural crossover.} Attention's marginal share $\beta P/(c+\beta P)$ reaches 50\% at 124K tokens for Qwen3.6-27B and 30K for its synthetic full-attention counterpart, which replaces the recurrent layers by full attention and recomputes both coefficients.}
\label{fig:econ-pstar-curves}
\end{figure}

At $w=1$, a position with $P$ available keys costs $c+\beta P$; the two components are equal at $P=P^*$. Token processing dominates below this context length and attention above it. Figure~\ref{fig:econ-pstar-curves} and the architecture map (Figure~\ref{fig:econ-architectures}, Appendix~\ref{app:econ-coefficients}) show how the rate varies across models.

Serving sets the rate for thinking again. Suppose a token the model wrote will be needed once more and is not returned in later prompts. With an unbroken reusable prefix, keeping it costs a linear rereading charge $(1-\rho)c$; generating it again costs $wc$. The original generation is common and cancels. The extra linear cost of regeneration, in attended pairs, is

\begin{equation}
 P^*_{\rm regen}=\frac{wc-(1-\rho)c}{\beta}=(w+\rho-1)P^*.
 \label{eq:econ-regeneration-rate}
\end{equation}

Retaining $r$ tokens through $L$ later evaluated positions incurs carry $\beta rL$; regenerating them once instead costs an extra $\beta rP^*_{\rm regen}$ in linear work. Ignoring attention during regeneration or rereading, keeping is cheaper while $L<P^*_{\rm regen}$: the exchange rate, moved by the serving, is also the horizon past which carrying a token costs more than writing it again. This compares two ways of producing the same token; the retention horizon below holds the continuation fixed, so nothing is regenerated, and asks when carrying a block costs more than rebuilding what followed it.

\subsection{Retaining and editing history}
\label{sec:econ-editing}

Fix the generated text and external inputs, and write $\Delta M$ for the cost of handling this continuation one way minus the cost of handling it another; in the retention comparisons the first way is Keep and the second Drop. For a block of $n$ input positions following a reusable prefix of length $H$, the rebuilding cost is $A(H,n)=cn+\beta[Hn+n(n+1)/2]$: processing the rebuilt tokens and their attention to the prefix and to one another. Let $L$ count the later positions evaluated by both arms, repeats included, excluding any rebuild or production that only one arm performs.

\paragraph{A retention horizon.}
Suppose $r>0$ reasoning tokens follow a prefix of $H$ tokens and precede an answer of $a$ tokens. With generated-state reuse, keeping the block charges every one of the $L$ later evaluations $\beta r$ for attending to it, while dropping it breaks the cache at the block and rebuilds the answer once:

\begin{align}
 \Delta M_{\rm keep}&=\beta rL-A(H,a),\label{eq:econ-retention}\\
 L^*&=\frac{a}{r}\left(P^*+H+\frac{a+1}{2}\right).\label{eq:econ-horizon}
\end{align}

Setting $\Delta M_{\rm keep}=0$ gives $L^*=A(H,a)/(\beta r)$, and $c/\beta=P^*$ turns it into Equation~\ref{eq:econ-horizon}. Keeping is mechanically cheaper for $L<L^*$: the horizon is the amount of later work at which the accumulated carry equals the answer's one-time rebuild. Under prompt-only reuse, both arms rebuild the answer, so those evaluations belong to $L$, and keeping also rereads the reasoning,

\begin{equation}
 \Delta M_{\rm keep}=\beta rL+A(H,r)>0,
 \label{eq:econ-retention-zero}
\end{equation}

so keeping is never mechanically cheaper there. Both comparisons assume at least one later call, an exact reusable prefix and no subsequent edit or eviction; the initial generation cancels (Appendix~\ref{app:econ-derivations}).

\paragraph{An edit.}
The same account prices any client edit. Start with a fully cached history. Replace $b$ tokens after prefix $H$ with $s<b$ tokens, leaving $\ell$ later history tokens in place. If only $H$ remains reusable, the next call rebuilds the replacement and the surviving tail. Let $K_{\rm gen}$ be the cost of producing the replacement, charged once (Appendix~\ref{app:econ-derivations} prices a separate production call). With $L$ as above, the edited-minus-unchanged cost is

\begin{equation}
 \Delta M_{\rm edit}=K_{\rm gen}+A(H,s+\ell)-\beta(b-s)L.
 \label{eq:econ-edit}
\end{equation}

The edit pays production and rebuilding to save attention to $b-s$ keys at every later evaluation, and is mechanically cheaper when the saving exceeds the first two terms. Its location matters as well as its size: a long surviving tail makes the same replacement more expensive, and at fixed $b,s,\ell,L$ each additional prefix token raises rebuilding cost by $\beta(s+\ell)$. Appendix~\ref{app:econ-derivations} states the cache assumptions; Equation~\ref{eq:econ-cost} prices other cache transitions through their actual reusable prefixes.

\subsection{Mechanical and behavioral change}
\label{sec:econ-split}

Changing history also changes the continuation: Figure~\ref{fig:econ-traces} (Appendix~\ref{app:econ-trace-sources}) shows reasoning and actions repeated under Keep and a plan restated under Drop. We now price this changed work alongside the mechanical effect.

Let policy 1 and baseline 0 produce continuations $\xi_1$ and $\xi_0$ at one architecture, cache rule and decoding premium. Their total cost difference is $\Delta C=C(1;\xi_1)-C(0;\xi_0)$. Replay $\xi_1$ under both policies, with rendered prompts, cache transitions, replacement text and any calls producing it specified. Adding and subtracting $C(0;\xi_1)$ gives

\begin{align}
 \Delta M&=C(1;\xi_1)-C(0;\xi_1),\nonumber\\
 \Delta B&=C(0;\xi_1)-C(0;\xi_0),\nonumber\\
 \Delta C&=\Delta M+\Delta B.
 \label{eq:econ-decomposition}
\end{align}

The mechanical term changes how the same continuation is processed; replay computes it. The behavioral term changes the continuation while the processing policy stays fixed; rerunning supplies the new continuation and the account prices it under the baseline. Positive terms increase cost, so $-\Delta B$ is a behavioral saving. Policy 1 is Keep in the retention and compaction comparisons and the client window in the window replay. Policy 1 costs less exactly when

\begin{equation}
 \boxed{-\Delta B>\Delta M.}
 \label{eq:econ-repay}
\end{equation}

A mechanical expense requires a larger behavioral saving; a mechanical saving can absorb some increase in later work. The chosen continuation reference determines the allocation between $\Delta M$ and $\Delta B$; reversing it preserves their sum. Appendix~\ref{app:econ-derivations} gives both allocations and the decomposition in expectation.

The retention rule must identify which reasoning it removes. Our editing experiment (Section~\ref{sec:econ-repayment}) retains visible \texttt{THOUGHT} text under both native-reasoning policies. Appendix~\ref{app:econ-thought} explains the two channels and their retention rules.

\subsection{Changing the system}
\label{sec:econ-decisions}

The same recordings can be priced under different architectures and serving rules. For two arms whose evaluated-position, attended-pair and generation counts differ by $\Delta N_\rho$, $\Delta Q_\rho$ and $\Delta G$, dividing their cost difference by $\beta$ gives

\begin{equation}
 \frac{\Delta C}{\beta}=P^*\{\Delta N_\rho+(w-1)\Delta G\}+\Delta Q_\rho.
 \label{eq:econ-ranking-surface}
\end{equation}

Its zero locates the prices at which the policies exchange rankings (Appendix~\ref{app:econ-derivations}). Cache rules change the evaluated-token and attended-pair counts; architecture changes $P^*$; the decoding premium changes the price of excess generation. These changes reprice both the mechanical and behavioral terms, even though the recorded reasoning and actions stay fixed. Moving from $w=1$ to five changes the cost difference by $4c\Delta G$, so any deletion, summary or edit that induces extra generation carries a greater downside at the higher premium. These are prices on recorded work: changing the model, task, harness or supplied history also changes which continuations occur, and the live comparisons of Section~\ref{sec:econ-experiments} measure that response, with native success evaluated alongside cost (Appendix~\ref{app:econ-derivations}).

\begin{figure*}[t]
\centering
\includegraphics[width=\textwidth]{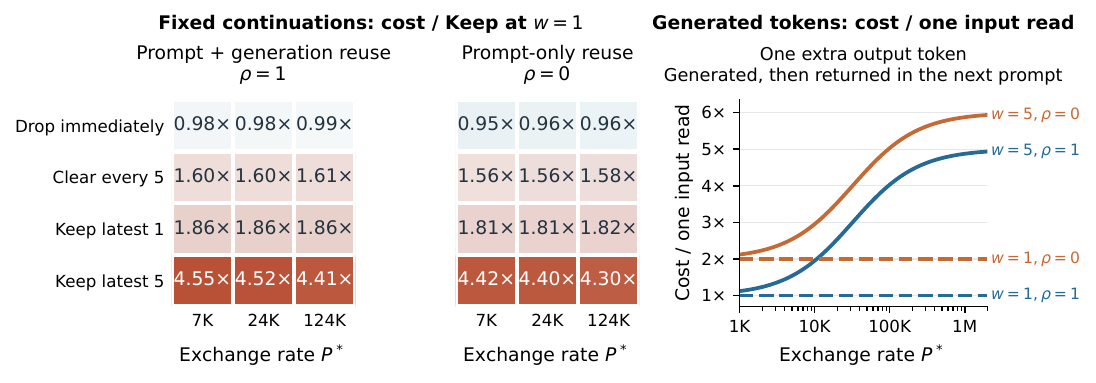}
\caption{\textbf{Editing history and generating more work have different costs.} Left: schedule/Keep cost on 132 fixed continuations ($w=1$, 16-token blocks) at the Qwen3-30B-A3B, Qwen3-8B and Qwen3.6-27B exchange rates. Right: cost of generating a token and returning it in the next prompt, relative to one input read, at 32K keys.}
\label{fig:econ-mechanical-policies}
\end{figure*}

\begin{figure*}[t]
\centering
\includegraphics[width=.85\textwidth]{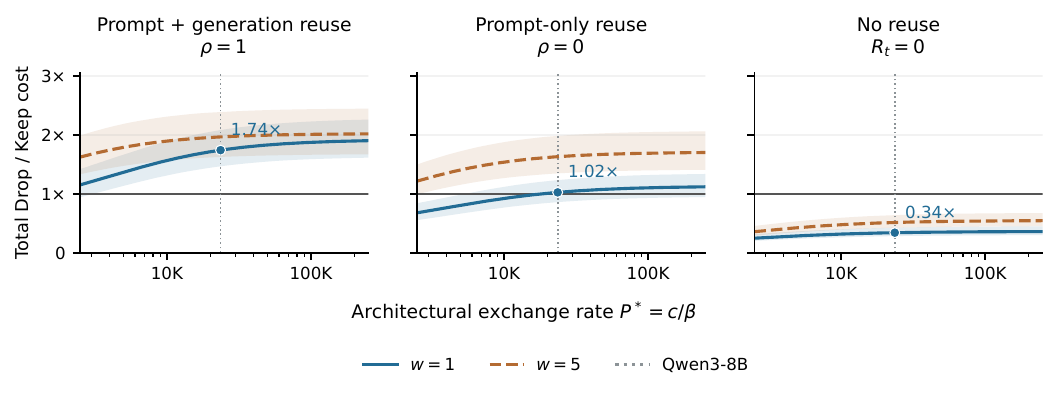}
\caption{\textbf{Prices reverse rankings on unchanged work.} Total Drop/Keep cost of the same 103 Qwen3-8B ScienceWorld pairs under three cache rules, with pointwise 95\% family-bootstrap bands; the dotted line marks the 8B exchange rate.}
\label{fig:econ-repricing}
\end{figure*}

\begin{figure*}[t]
\centering
\includegraphics[width=.9\textwidth]{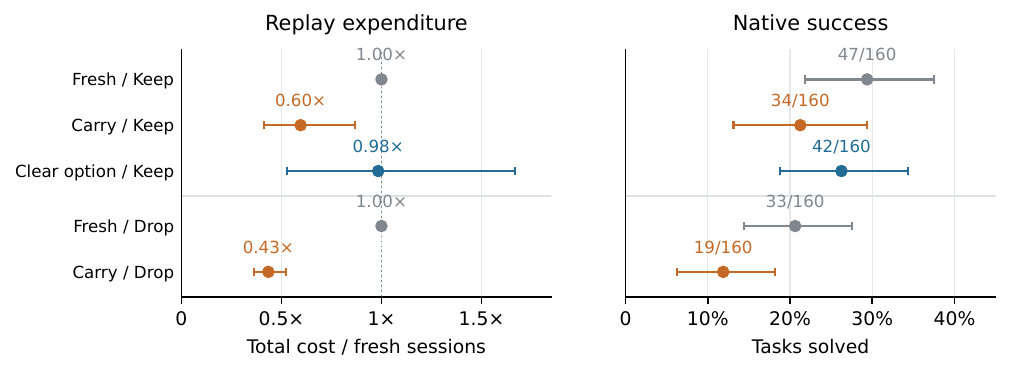}
\caption{\textbf{Lower cost comes with fewer completed tasks.} Qwen3-8B on 32 matched five-task chains, 160 task executions per arm. Cost is replay cost at $\rho=1$, $w=1$, relative to Fresh under the same retention rule; bars are 95\% chain-bootstrap intervals.}
\label{fig:econ-chained-history}
\end{figure*}

\begin{figure}[!ht]
\centering
\includegraphics[width=\columnwidth]{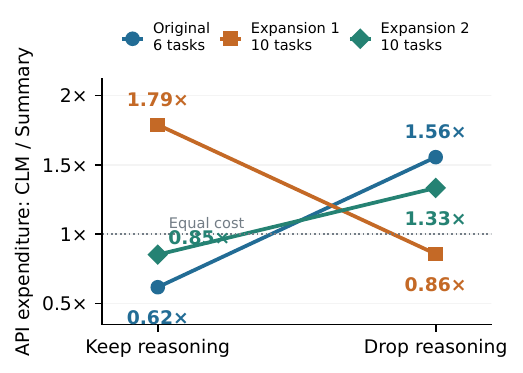}
\caption{\textbf{Client-policy rankings change across cohorts.} CLM/Summary ratio of total API cost in dollars, compaction included; below one, CLM costs less. Five matched task streams yield 20 complete chains: four on the original six-task stream and eight on each expansion's two five-task streams. Appendix~\ref{sec:econ-client-factorial} gives outcomes and configurations.}
\label{fig:econ-compaction}
\end{figure}
\section{Related work}\label{sec:econ-related}

\paragraph{Pricing context management.}
Token reduction does not imply a smaller cost: \citet{satish2026tokensavingssystematicstudy} compare 65 policies across three models, and \citet{hoy2026continuouscontextmanagement} report that per-call rewrites reduce input tokens while increasing estimated API costs for Opus and Kimi under provider caching. Existing break-evens compare future token savings with summary production \citep{arxiv260623525,wang2026statecomplearningcompresshistory}; other evaluations price cached input with a flat discount \citep{zhang2026autocompactlearningcompactcontext}.

\paragraph{Cache-aware interventions.}
History edits invalidate prefix reuse from the first changed position \citep{wang2026statecomplearningcompresshistory}. ACON discusses rebuilding as a limitation but excludes input caching from its API-cost calculation \citep{arxiv251000615}; Beyond Compaction discusses repeated invalidation and cache-write overhead \citep{arxiv260611213}. CliffCompaction measures sliding-window re-prefill costs \citep{nguyen2026cliffcompactioncostefficientcompactionlonghorizon}. KV-streams avoids repeated compaction prefill by carrying retained KV entries across compactions \citep{penaloza2026kvstreamsefficientcompactionagentic}. CLM counts prefix-reuse FLOPs call by call while allowing model-directed edits \citep{shao2026contextlanguagemodels}.

\paragraph{Reasoning and induced trajectories.}
SWE-AGILE uses supervised fine-tuning and reinforcement learning within a sliding reasoning window that retains older digests \citep{lian-etal-2026-swe}; state commitment learning trains erasure-compatible reasoning \citep{ding2026statecommitmentlearningtraining}; Memento masks completed reasoning inside the engine \citep{arxiv260409852}. When Can Agents Forget studies trajectory amplification after deletion \citep{wang2026agentsforgetreasoningiclr}; Plans Fade Fast identifies a reasoning-trace confound in plan stripping \citep{mehta26plans}. Our comparisons combine these concerns: they price rebuilding and retained-context carry, then measure how changes in the continuation alter the total. Appendix~\ref{app:econ-hosted} also records earlier hosted collections; provider retention rules appear in the same appendix.
\section{Experimental evidence}
\label{sec:econ-experiments}

Cost ratios divide total cost over matched populations, including native failures. ScienceWorld and HotpotQA intervals resample task families or questions \citep{wang2022scienceworld,yang2018hotpotqa}; sequential-task intervals resample whole chains. Appendix~\ref{app:econ-protocol} specifies populations and accounting conventions.

\subsection{Predicted computation versus GPU execution}

We predict computation for prescribed token sequences, then execute them through instrumented vLLM \citep{kwon23vllm} on an RTX 3070. Operation counts agree with predictions in all 648 requests (Figure~\ref{fig:econ-validation-parity} and Table~\ref{tab:econ-validation}, Appendix~\ref{app:econ-accounting}). Appendix~\ref{app:econ-accounting} details the operators, cache checks and final-token convention. Operation counts do not give latency, because measured decode time also depends on batch size (Appendix~\ref{app:econ-timing}), so we sweep the generation premium $w$ instead of fixing it.

\subsection{Shorter history, more computation}
\label{sec:econ-mechanical-policies}

A sliding reasoning window removes an older block whenever it advances. SWE-AGILE combines this schedule with retained digests \citep{lian-etal-2026-swe}. Our replay applies only its deletion schedule, without the digests or the trained behavior. We isolate deletion timing on 132 hosted ScienceWorld recordings, preserving every generated response, action and external input (Figure~\ref{fig:econ-mechanical-policies}).

Across $P^*\in[1\mathrm{K},2\mathrm{M}]$ and $w\in[1,5]$, keeping the latest five blocks costs 3.97--4.56 times Keep with generated-state reuse, despite shortening submitted prompts by 1.6\%. Each advancing window rebuilds surviving history. Batching reduces this expense; with generated-state reuse, immediate deletion stays within 2\% of Keep throughout the sweep (Table~\ref{tab:econ-mechanical-policies}). Without prefix reuse, all four deletion schedules are cheaper than Keep.

All schedules generate the same output, so the decoding premium cancels from their absolute differences. Figure~\ref{fig:econ-mechanical-policies} separately prices additional generation: at large $P^*$, producing and returning a token costs nearly five input reads under $w=5,\rho=1$, and six under $\rho=0$. Appendix~\ref{app:econ-mechanical-policies} gives the replay and this marginal account.

\subsection{Serving prices change the comparison}
\label{sec:econ-repricing}

Figure~\ref{fig:econ-cache-cost} (Appendix~\ref{app:econ-retention-share}) holds the continuation fixed. Generated-state reuse charges Keep for attention and Drop for rebuilding the following text. Prompt-only reuse also reprocesses Keep's previous reasoning, raising its cost while barely changing Drop's. Appendix~\ref{app:econ-retention-share} gives the replay and uncertainty.

We then price both live continuations from 103 matched forks (Figure~\ref{fig:econ-repricing}). Drop is more expensive with generated-state reuse and cheaper with no reuse; both intervals exclude equality (Table~\ref{tab:econ-repricing-uncertainty}). Prompt-only reuse lies near equality at $w=1$: keeping is never mechanically cheaper there (Equation~\ref{eq:econ-retention-zero}), but Drop's continuation is longer. Increasing the decoding premium favors Keep because Drop generates more output (Equation~\ref{eq:econ-ranking-surface}). Serving prices reverse the ranking while recorded responses and outcomes remain fixed.

\subsection{Retention changes what the agent solves}
\label{sec:econ-live}

Table~\ref{tab:econ-live} runs Keep and Drop live from the same forks at fixed serving, on five models and two environments. On ScienceWorld, dropping earlier reasoning costs three of the four Qwen3 models 1.7 to 1.9 times Keep's cost, with success differences that span zero. The exception is gpt-oss-20b, which solves 20 of 88 tasks under Drop against 3 under Keep in our custom Harmony histories (Appendix~\ref{app:econ-protocol}). On HotpotQA, with paragraph titles hidden and only the answer graded, Drop's cost ratios are 0.70 to 0.92, with intervals that include 1, and Drop loses 9 to 23 points of fully correct answers on all four models; a repeat 8B collection on the same questions reproduces the loss (Table~\ref{tab:econ-joint-outcomes}).

\begin{table}[t]
\centering\footnotesize\setlength{\tabcolsep}{1.6pt}
\begin{tabular}{@{}llrcc@{}}
\toprule
 & Model & $n$ & Cost D/K & Solved D$-$K (pp)\\
\midrule
SW & 8B & 103 & 1.74 [1.46, 2.08] & $-$3.9 [$-$10.2, 2.0]\\
SW & 4B & 126 & 1.18 [0.93, 1.49] & 0.8 [$-$8.8, 9.0]\\
SW & 30B-A3B & 119 & 1.73 [1.43, 2.09] & $-$0.8 [$-$6.6, 5.4]\\
SW & 32B & 52 & 1.94 [1.65, 2.26] & 0.0 [0.0, 0.0]\\
SW & gpt-oss & 88 & 1.20 [0.96, 1.49] & 19.3 [10.7, 29.3]\\
\midrule
HP & 8B & 165 & 0.81 [0.64, 1.00] & $-$13.3 [$-$20.6, $-$6.1]\\
HP & 4B & 122 & 0.70 [0.48, 1.04] & $-$13.9 [$-$22.1, $-$5.7]\\
HP & 30B-A3B & 88 & 0.92 [0.80, 1.06] & $-$22.7 [$-$33.0, $-$11.4]\\
HP & gpt-oss & 195 & 0.92 [0.78, 1.09] & $-$9.2 [$-$14.9, $-$4.1]\\
\bottomrule
\end{tabular}
\caption{Live Keep (K) and Drop (D) from the same forks, generated-state reuse, $w=1$. Cost: total D over total K. Solved: paired difference in full-score rate. 95\% cluster-bootstrap intervals. SW: ScienceWorld; HP: HotpotQA, titles hidden, answer-only grading. Qwen3 unless named. The 32B outcomes are identical.}
\label{tab:econ-live}
\end{table}

\subsection{Carrying history across task boundaries}
\label{sec:econ-sequential}

We run Qwen3-8B on 32 sequences of four distinct ScienceWorld tasks followed by a repeat of the first. Fresh or carried history is crossed with Keep or Drop reasoning; a fifth arm offers a \texttt{clear memory} action. Arm order is randomized within each chain, with up to ten calls per task.

Carrying history reduces cost and native success under both retention rules (Figure~\ref{fig:econ-chained-history}). Under Keep, carried history costs 0.60 of fresh sessions and solves 17 of 96 new tasks against 30; under Drop, it costs 0.43 and solves 13 against 26 (Table~\ref{tab:econ-chained-history}). The success loss is concentrated in subsequent new tasks. The clearing arm lies between Fresh and Carry in success, at an aggregate cost close to Fresh. Appendix~\ref{app:econ-sequential} (Table~\ref{tab:econ-chained-history}) reports outcomes by task position and the changes in generated reasoning.

\subsection{Summarizing history and editing it}
\label{sec:econ-repayment}

A client's cost depends on the server's retention. With generated-state reuse, a client window that re-truncates history at every call adds 352.9\% and 145.9\% of full-history cost under Keep and 45.3\% and 31.0\% under Drop, on 114 ScienceWorld and 171 HotpotQA forks, mostly mechanically (Appendix~\ref{app:econ-window-replay}). For a published method, we compare automatic Summary with the zero-shot CLM editing prompt \citep{shao2026contextlanguagemodels} on hosted Qwen3.7-Plus, crossed with Keep/Drop native reasoning. Following \citet[Sec.~6.2]{arxiv260611213}, tasks share history but receive fresh sandboxes and verifiers. Chaining exercises compaction across individually tractable tasks and tests whether agents clear obsolete history. We complete 20 chains: four on a six-task stream and sixteen on four new five-task streams. Summary triggers at 75\% of 32K; CLM edits through the shell.

In Figure~\ref{fig:econ-compaction} the ranking varies with retention and between cohorts, and CLM's API cost is 0.62 to 1.79 times Summary's. CLM costs less than Summary under Keep and more under Drop in the original and latest cohorts; the intermediate cohort reverses both comparisons. In the latest cohort, CLM is 15\% cheaper under Keep and 33\% dearer under Drop. Native passes are 9 versus 7 under Keep and 9 versus 9 under Drop. Each cell is one unrepeated chain per stream at temperature 0.7, so these reversals are not separated from run-to-run variation. Appendix~\ref{sec:econ-client-factorial} (Table~\ref{tab:econ-chain-totals}) reports the separate task streams, output allowances and harness revisions.

In the original cohort's prompt-only replay, CLM's shorter Keep continuation more than repays its mechanical expense. The first expansion includes 11 automatic CLM Keep history rollbacks; the latest has none. All eight latest chains compact and continue. Both the client and its execution setting enter the comparison.
\FloatBarrier
\section{Discussion and conclusion}

The cost model prices a run at two architectural rates plus the serving's cache rule and decoding premium, and its counts matched an instrumented engine on all 648 requests. Repeated rebuilding can dominate a run's cost while barely shortening history. With generated-state reuse, retention and immediate deletion become mechanically close on the hosted recordings and on the 8B and 32B panels, but not on 4B and 30B-A3B, where retention adds 38.6\% and 27.4\% of Keep cost (Table~\ref{tab:econ-reference-sensitivity}).

The two parties have complementary choices. A server can preserve generated states and expose which prefixes remain reusable, and a client can then price an edit's replacement and surviving tail before choosing where and when to cut. A reward calibrated to one cache rule and decoding price embeds that deployment choice in the editor it trains.

Even economical edits change the work that follows. Carrying history reduces cost while solving fewer new tasks, and dropping earlier reasoning costs HotpotQA answers on all four models in Table~\ref{tab:econ-live}. This gives a practical order for transferring a policy: reprice its recorded work under the destination system, then evaluate the continuation and success it produces there.

\section*{Limitations}

\paragraph{Cost measure.} Costs are operation counts at declared coefficients, except in the compaction study, which uses the provider's billed tariff. Elapsed time, memory traffic and energy need not follow the counts, because decoding is limited by memory traffic. Appendix~\ref{app:econ-timing} measures time for two models on one GPU, and the decoding premium $w$ is swept over a range rather than calibrated to a deployment. The operation counts were verified in one engine, on a dense and a hybrid model. Replays reconstruct cache states from the logged prompts instead of reading them from the server. Hosted recordings are priced at the coefficients of a reference open model. The split between mechanical and behavioral change depends on which continuation serves as the replay reference. Appendix~\ref{app:econ-derivations} gives both allocations, and their sum does not depend on that choice.

\paragraph{Experiment coverage.} In the open-model fork and chain studies, episodes run to 30 calls, 12 on HotpotQA and 10 per chained task, within a 32K window, or 16K for the Qwen models on HotpotQA; the hosted recordings reach 66 calls and the compaction chains 207. Work beyond the horizon $L^*$ of Equation~\ref{eq:econ-horizon} is reached by repricing recorded work. The keep-or-drop rule is not evaluated as a deployed policy, and its arms in the fork collections appear only in Tables~\ref{tab:econ-terminal-outcomes} and~\ref{tab:econ-inventory}. The open-model collections cover Qwen3 models from 4B to 32B, gpt-oss-20b and an R1 distill, on ScienceWorld and on HotpotQA, where only the answer is graded. The models were served at different precisions, so differences between models in Table~\ref{tab:econ-live} are not architectural effects. Appendix~\ref{app:econ-controls} lists every collection, its controls and its full outcomes. The compaction study uses CLM's released zero-shot prompt on a hosted model. It covers 20 chains in three cohorts with different harness revisions and output allowances, sampled at temperature 0.7 with one execution per cell. Provider retention rules are taken from documentation that changes between model versions.

\paragraph{Fork collection design.} The open-model fork collections use fixed arm order and action-replay resets without a state-equivalence check. They condition on reaching the fork and completing the planned arms; native failures remain in that population. The chained-task study randomizes arm order. The engine is not deterministic at temperature zero, and a second Keep execution from the same fork differs from the first (Table~\ref{tab:econ-inventory}), so comparisons are paired within one collection.

\section*{Ethics Statement}

\paragraph{Use of AI assistants.} We used AI coding and writing assistants, Claude (Anthropic) and Codex (OpenAI).

\section*{Acknowledgments}

We thank Iv\'an Belenky and Tamarillo Labs for the discussions that shaped our ideas on the behavioral effects of context policies, and for lending the token resources that funded the experiments.
\bibliography{anthology,custom}
\clearpage
\appendix
\makeatletter
\renewenvironment{figure}{\@dblfloat{figure}}{\end@dblfloat}
\renewenvironment{table}{\@dblfloat{table}}{\end@dblfloat}
\makeatother
\section{Accounting and decision rules}
\label{app:econ-derivations}

This appendix derives the equations used in the body. The counted object is a specified sequence of evaluations, together with its cache transitions. A replay specifies which evaluations a fixed continuation requires under a cache rule; a live run supplies the continuation produced by the agent.

\subsection{The accounting contract}

Fix a tokenizer, rendered prompts, an initial cache, and a serving rule. At call $t$, let $P_t$ be prompt length, $R_t$ its actually reusable contiguous prefix, and $G_t$ the number of subsequent evaluated generation positions. The admissible range is $0\leq R_t\leq P_t$. The same token evaluated twice counts twice. A cached key costs no new token evaluation but remains a key for later queries. For the architectural prices $c,\beta>0$ and decoding premium $w>0$, write $\operatorname{tri}(n)=n(n+1)/2$. Rebuilding $n$ positions after an $H$-token reusable prefix costs
\begin{equation}
 A(H,n)=cn+\beta\{Hn+\operatorname{tri}(n)\}.
 \label{eq:econ-rebuild-function}
\end{equation}
It prices $n$ consecutive evaluated positions after an $H$-token prefix. It does not price producing new replacement text in a separate call.

\paragraph{Per-call identity.}
Prompt queries occupy positions $R_t+1,\ldots,P_t$, and generation queries occupy $P_t+1,\ldots,P_t+G_t$. Let $N_t$, $Q_t$ and $C_t$ denote this call's evaluated positions, attended pairs and cost. Summing positions gives
\begin{align}
 N_t&=P_t-R_t+G_t,\nonumber\\
 Q_t&=\operatorname{tri}(P_t+G_t)-\operatorname{tri}(R_t),\nonumber\\
 C_t&=cN_t+\beta Q_t+(w-1)cG_t.
 \label{eq:econ-app-call}
\end{align}
Summing over calls proves Equations~\ref{eq:econ-cost}--\ref{eq:econ-count-q}. The premium changes generation's linear price to $wc$; it does not multiply attention work.

\paragraph{Emitted tokens and evaluated positions.}
In an ordinary autoregressive request returning $O_t>0$ tokens, the first output is sampled from prefill logits. Only $O_t-1$ output tokens have subsequently passed through the transformer when the request ends. Thus executed decode work uses $G_t=O_t-1$; the final emitted token becomes an evaluation when later forwarded. The archived accounting-validation experiment uses this convention. The live fork study instead prices its recorded output count as $G_t=O_t$, following the producer's reduced-account convention. With the same $P_t,R_t$, that convention adds $wc+\beta(P_t+O_t)$ to the executed-position price of each nonempty request. The resulting one-position offset is included in the boundary-convention check in Appendix~\ref{app:econ-accounting}. Speculation, rejected tokens, and extra logit rows require their actual evaluation counts.

For a total, mechanical or behavioral cost difference $\Delta X$, with $X\in\{C,M,B\}$, let $\Delta N_X$, $\Delta Q_X$ and $\Delta G_X$ be the count differences between the same two executions. Subtracting their accounts gives:
\begin{equation}
 \Delta X=c\Delta N_X+\beta\Delta Q_X+(w-1)c\Delta G_X.
 \label{eq:econ-priced-differences}
\end{equation}

\paragraph{What the two coefficients omit.}
Let $J$ count output-head rows, $\eta$ price one output-projection row, and $W_{\mathrm{state}}$ count recurrent-state and convolution work. The fuller operator count is
\begin{equation}
 F_{\mathrm{operators}}=cN+\beta Q+\eta J+W_{\mathrm{state}}.
 \label{eq:econ-app-full}
\end{equation}
The recurrent layers' projection matrices are already included in $c$; $W_{\mathrm{state}}$ adds their state updates. Head work cancels only if both compared executions have the same $J$. It cancels in the dense validation pairs; the hybrid pairs retain the measured head difference. For a per-token state-update price $\omega$, folding state work into a constant token price requires $W_{\mathrm{state}}=\omega N$ on the execution path in question. Chunked prefill and recurrent decoding need not satisfy that equality. The reduced prices omit the additional head and state-update terms, normalization, activation functions, and data movement. A deployment-specific latency model or API tariff supplies the corresponding time or dollar cost. Either can replace $C$ in the repayment algebra, provided all three terms use that same model.

\subsection{The call in Figure~\ref{fig:econ-cache-blocks}}
\label{app:econ-worked-cache}

The history in Figure~\ref{fig:econ-cache-blocks} is
\[
 [\underbrace{\text{task}}_{16{,}000}\mid
   \underbrace{\text{reasoning}}_{2{,}000}\mid
   \underbrace{\text{action}}_{100}\mid
   \underbrace{\text{new input}}_{500}],
\]
followed by 100 evaluated generation positions.
Drop removes only reasoning. Token-granular reuse produces the next-call accounts in Table~\ref{tab:econ-worked-cache}.

\begin{table}[ht]\centering\small\setlength{\tabcolsep}{4pt}
\begin{tabular}{@{}llrrrrrr@{}}\toprule
$\rho$ & Policy & $P$ & $R$ & $G$ & $N$ & $Q$ & $C$ (T) \\ \midrule
1 & Keep & 18,600 & 18,100 & 100 & 600 & 11,040,300 & 33.56 \\
1 & Drop & 16,600 & 16,000 & 100 & 700 & 11,445,350 & 38.59 \\
0 & Keep & 18,600 & 16,000 & 100 & 2,700 & 46,846,350 & 149.91 \\
0 & Drop & 16,600 & 16,000 & 100 & 700 & 11,445,350 & 38.59 \\
\bottomrule\end{tabular}
\caption{Next-call prices for the identical history under two reuse rules.}\label{tab:econ-worked-cache}
\end{table}

At $\rho=1$, Keep avoids rebuilding the action and costs less on this call. At $\rho=0$, both arms rebuild the action, and Keep also rebuilds the reasoning. Both calculations use the same subsequent input and generation, isolating the cost of retaining the thought under each cache rule.

\subsection{Repayment, expectation, and the replay reference}

Write $C_i(\xi)=C(\pi_i;\xi)$ at one fixed configuration. A cross-policy replay is defined only when the counterfactual rendering, cache transitions, replacement text, and any extra production calls are specified. Let $\xi_1$ and $\xi_0$ be the recorded continuations. Then
\begin{equation}
\begin{split}
 C_1(\xi_1)-C_0(\xi_0)
 &=\underbrace{C_1(\xi_1)-C_0(\xi_1)}_{\Delta M_1}\\
 &\quad+\underbrace{C_0(\xi_1)-C_0(\xi_0)}_{\Delta B_1}.
\end{split}
 \label{eq:econ-app-repay}
\end{equation}
\emph{Proof.} Add and subtract $C_0(\xi_1)$. The intervention costs less exactly when $-\Delta B_1>\Delta M_1$. This includes negative mechanical changes: then some increase in continuation cost is affordable. Each difference inherits the linear combination of $N,Q,G$ in Equation~\ref{eq:econ-priced-differences}.\hfill$\square$

The alternative reference gives
\begin{align}
 \Delta M_0&=C_1(\xi_0)-C_0(\xi_0),\nonumber\\
 \Delta B_0&=C_1(\xi_1)-C_1(\xi_0),\nonumber\\
 \Delta M_1-\Delta M_0&=-(\Delta B_1-\Delta B_0).
 \label{eq:econ-app-reference}
\end{align}
Both allocations sum to the same total. Their difference is the interaction between replay policy and continuation, so each component is reported with its reference. The retention replay uses Keep's continuation; the client-window replay uses the windowed continuation.

For a common task population, let $\mathbb P_i$ be the continuation distribution under policy $i$ and $\mathbb E_i$ its expectation. Assuming finite expectations and defined cross-policy replays,
\begin{align}
 \mathbb E_1[C_1]-\mathbb E_0[C_0]
 &=\underbrace{\mathbb E_1[C_1-C_0]}_{\text{mechanical}}\nonumber\\
 &\quad+\underbrace{\mathbb E_1[C_0]-\mathbb E_0[C_0]}_{\text{continuation}}.
 \label{eq:econ-app-expectation}
\end{align}
The first term is the mechanical change on policy 1's continuations, and the second is the continuation change at the baseline price.
\emph{Proof.} Add and subtract $\mathbb E_1[C_0]$. No equality of continuation distributions is assumed.\hfill$\square$
An expected-cost decision minimizes $\mathbb E_i[C_i]$ subject to a chosen minimum success probability $q$, for example $\Pr_i(\mathrm{success})\geq q$, where $\Pr_i$ denotes probability under policy $i$. An observed success count is an estimate of this constraint, not proof of success equivalence. The collection-specific intervals in Appendix~\ref{app:econ-controls} do not remove selection into a completed fork panel.

\subsection{Editing a specified history}

Start with a fully reusable history $[A\mid B\mid T]$, with lengths $H,b,\ell$. Replace $B$ by specified text $S$ of length $s<b$. Assume the new longest reusable prefix is exactly $A$, token-granular reuse, no alternative cached branch, no eviction or capacity event, and an append-only continuation after the edit. Let $L$ count later evaluated positions common to the two replays, excluding the edited arm's $s+\ell$ rebuilt positions and any separate production call. All common queries attend to the whole retained history.

\paragraph{Edit identity.}
Let $K_{\mathrm{gen}}$ be the incremental replacement-production cost, charged once. The edit-minus-unchanged cost is Equation~\ref{eq:econ-edit},
\begin{equation*}
 \Delta M_{\mathrm{edit}}=K_{\mathrm{gen}}+A(H,s+\ell)-\beta(b-s)L.
 \tag{\ref{eq:econ-edit}}\label{eq:econ-app-edit}
\end{equation*}
\emph{Proof.} Only the edited arm rebuilds $S,T$, paying $A(H,s+\ell)$. Each common query sees $b-s$ fewer keys, saving $\beta(b-s)$; its token-processing charge cancels. Charge replacement production once, as the incremental $K_{\mathrm{gen}}$. Sum the three terms.\hfill$\square$

For a separate call with a fully reusable $P_0$-token prompt that charges $s$ generation positions, Equation~\ref{eq:econ-cost} gives $K_{\mathrm{gen}}=wcs+\beta\{P_0s+\operatorname{tri}(s)\}$. Charging $s$ positions for $s$ emitted tokens uses the output-count convention above. Other production calls must use their actual reusable prefixes and generation counts in Equation~\ref{eq:econ-cost}, including any additional prompt work. A replacement written inside an ordinary call is priced by that call's own positions.

The fixed-continuation decision is $\beta(b-s)L>K_{\mathrm{gen}}+A(H,s+\ell)$. If it fails, the continuation must save more than the nonnegative remainder. Holding $H,\ell,L$ and the cache boundary fixed, one extra replacement token raises rebuilding and carry by
\begin{equation}
\begin{split}
 &A(H,s+\ell+1)-A(H,s+\ell)+\beta L\\
 &\quad=c+\beta(H+s+\ell+1+L).
\end{split}
 \label{eq:econ-app-marginal-summary}
\end{equation}
Its additional generation price is separate. Appending a generated position after length $P$ costs $wc+\beta(P+1)$ in the evaluated-position convention. An insertion before other generated positions also changes their attention costs.

\paragraph{Reasoning retention as a special case.}
Generate $r>0$ reasoning tokens and $a\geq0$ visible-output tokens after $H$. Consider at least one reuse call, identical later inputs and outputs, no subsequent history edits or capacity events, and the same exact-prefix assumptions as above. The original generation is common and cancels. When $\rho=1$, Keep reuses all $H+r+a$ positions while Drop reuses exactly $H$. Every common later evaluation sees $r$ extra keys under Keep; only Drop rebuilds the visible output. Hence Equations~\ref{eq:econ-retention} and~\ref{eq:econ-horizon},
\begin{align*}
 \Delta M_{\mathrm{keep}}&=\beta rL-A(H,a),\tag{\ref{eq:econ-retention}}\\
 L^*&=\frac{a}{r}\left(P^*+H+\frac{a+1}{2}\right).\tag{\ref{eq:econ-horizon}}\label{eq:econ-app-retention}
\end{align*}
For $\rho=0$, both first reuse calls cache exactly $H$. The visible output is rebuilt in both and therefore belongs to $L$; only Keep rebuilds the reasoning. Consequently Equation~\ref{eq:econ-retention-zero},
\begin{equation*}
 \Delta M_{\mathrm{keep}}=\beta rL+A(H,r)>0.
 \tag{\ref{eq:econ-retention-zero}}\label{eq:econ-app-retention-zero}
\end{equation*}
\emph{Proof.} Sum $\beta r$ over common evaluations, then subtract Drop's unmatched answer rebuild or add Keep's unmatched reasoning rebuild, respectively. Subsequently those rebuilt prompt states stay cached. The shared generation premium cancels. Setting $\Delta M_{\mathrm{keep}}=0$ in the first case gives $L^*=A(H,a)/(\beta r)$. For $a>0$, substituting $c/\beta=P^*$ and $\operatorname{tri}(a)/a=(a+1)/2$ gives the displayed horizon. For $a=0$, $A(H,a)=0$ and the same horizon formula gives $L^*=0$.\hfill$\square$
Here $L$ counts evaluated positions, including repeated evaluations, not calls or unique future tokens. Including the unmatched rebuild in $L$ would charge its attention twice. With no subsequent reuse there is no future cost difference. Deleting an empty block is also a different geometry: it does not break an answer's cache.

\subsection{Changes of price}

\paragraph{Linear exchange-rate comparisons.}
Equation~\ref{eq:econ-regeneration-rate} subtracts one linear rereading charge $(1-\rho)c$ from one regeneration charge $wc$ and divides by $\beta$. Generated-state reuse gives $wP^*$; prompt-only reuse gives $(w-1)P^*$. Attention during rereading or regeneration and any cache rebuilding must be added for a complete comparison.

For a generated token that returns unchanged, the linear write-and-return price is $(w+1-\rho)c$. Relative to one input read at $c$, its spread is $(w-\rho)c$, zero at $w=\rho=1$. Appendix~\ref{app:econ-mechanical-policies} includes attention in the corresponding comparison used in Figure~\ref{fig:econ-mechanical-policies}.

\paragraph{Repricing fixed continuations.}

For fixed recorded evaluations, Equation~\ref{eq:econ-priced-differences} with $X=C$ reads
\begin{equation*}
 \Delta C(w)=c\Delta N+\beta\Delta Q+(w-1)c\Delta G.
\end{equation*}
Its slope in $w$ is $c\Delta G$. For $\Delta G\ne0$, its zero is
\begin{equation}
 w_{\mathrm{break}}=1-\frac{c\Delta N+\beta\Delta Q}{c\Delta G}.
 \label{eq:econ-wbreak}
\end{equation}
Dividing by $\beta$ proves Equation~\ref{eq:econ-ranking-surface}. A zero outside the tested price range produces no reversal inside it. The same derivation applies to sums or expectations of costs over a fixed population. It does not turn a median of individual cost ratios into a ratio of expected costs. Changing model behavior, rendering, or the sampling population requires new continuations, rather than a new price on the old ones.
\section{Architecture prices and serving configurations}
\label{app:econ-coefficients}

\begin{figure*}[t]
\centering
\includegraphics[width=.94\textwidth]{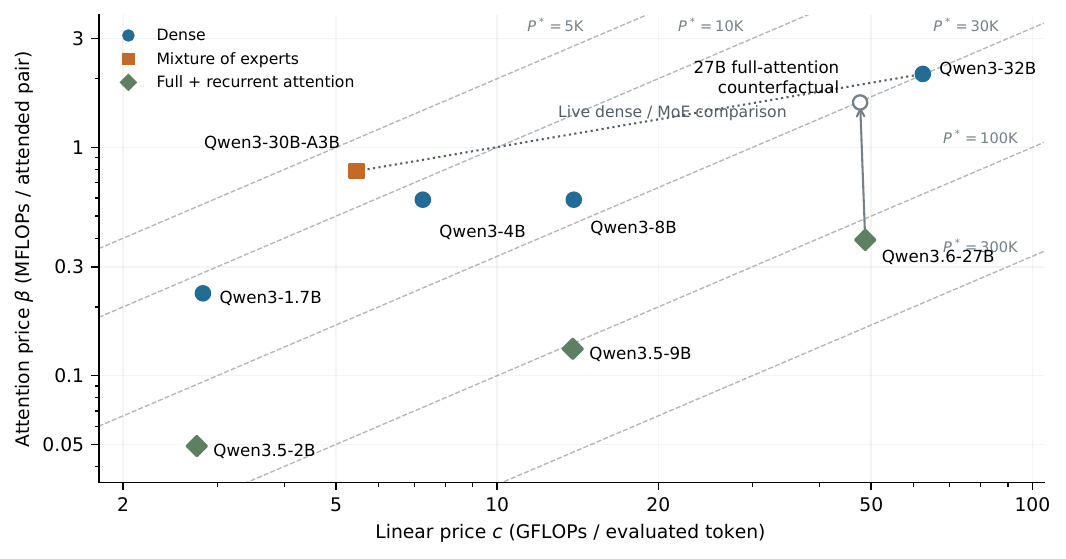}
\caption{\textbf{Architectures occupy different price regimes.} A point specifies linear price $c$ and full-attention price $\beta$; diagonals have constant $P^*=c/\beta$. Filled points include the open models used in the live or replay studies and the hybrid references. The arrow replaces recurrent layers of the 27B hybrid with full attention and recomputes both coefficients. That synthetic counterpart is distinct from the live dense/MoE comparison linked by the dotted segment. Appendix~\ref{app:econ-coefficients} gives the coefficient scope and values.}
\label{fig:econ-architectures}
\end{figure*}

Table~\ref{tab:econ-coefficients} lists the reduced price coefficients used in the figures and comparisons. Live-run prices use twice the rounded active body-parameter count. The four reference architectures and the synthetic counterpart use the projection counts below. The coefficients price arithmetic operations; the timing measurements below examine their relation to elapsed time.

\begin{table}[ht]\centering\small\setlength{\tabcolsep}{4pt}
\begin{tabular}{@{}lrrr@{}}\toprule
Model or counterfactual & $c$ (G/token) & $\beta$ (M/pair) & $P^*$ (K tokens) \\ \midrule
Qwen3-4B & 7.260 & 0.589824 & 12.31 \\
Qwen3-8B & 13.900 & 0.589824 & 23.57 \\
Qwen3-30B-A3B & 5.462 & 0.786432 & 6.95 \\
Qwen3-32B & 62.400 & 2.097152 & 29.75 \\
Qwen3-1.7B & 2.819 & 0.229376 & 12.29 \\
Qwen3.5-2B & 2.745 & 0.049152 & 55.86 \\
Qwen3.5-9B & 13.837 & 0.131072 & 105.57 \\
Qwen3.6-27B & 48.701 & 0.393216 & 123.85 \\
27B full-attention counterfactual & 47.647 & 1.572864 & 30.29 \\
gpt-oss-20b & 6.060 & 0.196608 & 30.82 \\
R1-Distill-Llama-8B & 13.960 & 0.524288 & 26.63 \\
\bottomrule\end{tabular}
\caption{Declared reduced coefficients. G and M denote decimal billions and millions; K denotes thousands.}\label{tab:econ-coefficients}
\end{table}

\paragraph{Coefficient construction.}
The live-run convention sets $c$ to twice the active body projection-parameter count, excluding embeddings and the output head. For full-attention layers, $\beta=4\sum_{\text{layers}} h_qd_h$, charging query--key and attention--value contractions with two operations per multiply-add. Grouped-query attention reduces the projection weights but does not replace the query width by the KV width in this pair coefficient. The hybrid config recipe separately counts MLP, full-attention projections, and gated recurrent-layer projections; recurrent-state updates are excluded as in Equation~\ref{eq:econ-app-full}. Table~\ref{tab:econ-projection-components} supplies the reference projection subtotals.

For each reference architecture, we count two operations per multiply-add in its projection matrices. With model width $d$, MLP width $d_f$, layer counts $n,n_f,n_r$, attention query/KV widths $u=h_qd_h$, $v=h_{kv}d_h$, and output-gate indicator $g$, the components are
\begin{align*}
 c_{\rm MLP}&=6ndd_f,\\
 c_{\rm attn}&=2n_f d\{(2+g)u+2v\},\\
 c_{\rm recur}&=2n_r d\{2h_kd_k+3h_vd_v+2h_v\},\\
 c&=c_{\rm MLP}+c_{\rm attn}+c_{\rm recur},\qquad \beta=4n_f u.
\end{align*}
The recurrent head counts $h_k,h_v$ and widths $d_k,d_v$ refer only to its projection matrices. The recurrent state update is outside this reduced account. For the hybrid references, this calculation uses the projection-only convention in \citet[Appendix~C, Eqs.~7--8]{shao2026contextlanguagemodels}; Table~\ref{tab:econ-projection-components} recomputes the subtotals from each model's dimensions.

\begin{table}[ht]\centering\small\setlength{\tabcolsep}{4pt}
\begin{tabular}{@{}lrrr@{}}\toprule
Reference & MLP & Attention projections & Recurrent projections \\ \midrule
Qwen3-1.7B & 2.113929216 & 0.704643072 & 0.000000000 \\
Qwen3.5-2B & 1.811939328 & 0.176160768 & 0.757334016 \\
Qwen3.5-9B & 9.663676416 & 0.939524096 & 3.233808384 \\
Qwen3.6-27B & 34.225520640 & 3.355443200 & 11.120148480 \\
27B full-attention counterpart & 34.225520640 & 13.421772800 & 0.000000000 \\
\bottomrule\end{tabular}
\caption{Reference projection subtotals in billions of operations per evaluated token. Their sum is the reduced token price.}\label{tab:econ-projection-components}
\end{table}

The synthetic 27B counterpart keeps the reference dimensions and replaces its recurrent layers by full attention. Both $c$ and $\beta$ are recomputed. It is a pricing construction with no trained checkpoint or live continuations. The actual Qwen3-32B and Qwen3-30B-A3B points instead price two live deployments. Their layer counts, active sizes, query widths, precision, and eligible task panels differ; connecting them on the map does not isolate an architectural effect.

\paragraph{Checkpoint and precision provenance.}
Appendix~\ref{app:econ-protocol} lists the archived launchers and precisions (BF16 for Qwen3-4B, Qwen3-8B and the R1 distill; GPTQ-Int4 for Qwen3-30B-A3B; AWQ for Qwen3-32B). The gpt-oss launcher names \texttt{openai/gpt-oss-20b}; the retained plan does not establish a precision-matched comparison with Qwen, and exact checkpoint revision hashes are absent from the launch commands.

The gpt-oss reduced pair price does not include a full sliding-window-layer account. Its full-score outcomes do not depend on that price. Quantization and kernel choices also affect throughput and storage, which require deployment-specific measurements. The sweep $w\in[1,5]$ examines whether saved reading pays for additional writing when generation carries a serving premium. At $w=5$, generated tokens carry five times the linear price of prompt processing; Equation~\ref{eq:econ-wbreak} gives the resulting decision boundary.

\paragraph{Architecture and serving play different roles.}
The coefficients price evaluations. Serving selects which positions are evaluated and reused. Two servers running the same weights therefore use the same declared $(c,\beta)$ here but can produce different $(N,Q)$ through caching, rendering, and stopping rules. Repricing at another architecture's coefficients isolates the price change on the recorded continuation.

\subsection{Measured timing across batching regimes}
\label{app:econ-timing}

\begin{figure}[htbp]\centering
\includegraphics[width=.94\textwidth]{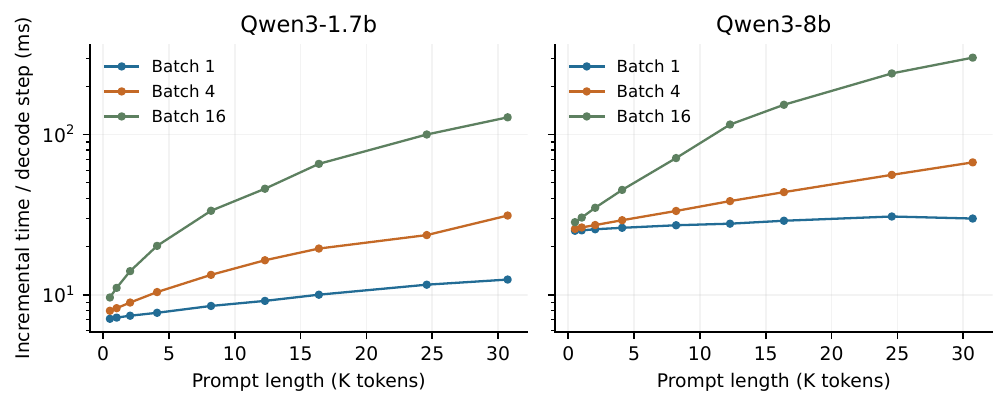}
\caption{\textbf{Serving changes the time price.} Measured incremental decode time under three batch sizes. The account's architectural coefficients remain fixed within each panel, while batching changes the time paid for an additional generation step. Each point uses the stated subtraction and minimum over three repetitions.}\label{fig:econ-timing}
\end{figure}

The timing controls measure Qwen3-1.7B and Qwen3-8B at prompt lengths from 512 to 30,720 tokens and batches of 1, 4 and 16. For each setting, the driver times requests producing one and 129 tokens, then estimates incremental decode time as $(T_{129}-T_1)/128$. Figure~\ref{fig:econ-timing} shows the minimum of three repetitions at each setting. These are end-to-end request timings; the subtraction removes the shared prefill contribution approximately.

Linear fits $t(P)=\alpha+\gamma P$ to these points give time crossovers $\alpha/\gamma$ of 39.0K, 10.1K and 1.15K tokens for 1.7B and 137.5K, 17.9K and 1.32K for 8B, in increasing batch-size order. The two batch-one crossovers extrapolate beyond the measured prompt range. These time crossovers are distinct from the architectural $P^*$: they depend on the serving workload. Their batch dependence motivates calibrating the generation price for the deployment workload.
\section{Experiment populations and configurations}
\label{app:econ-protocol}

The forked open-model analysis uses per-call records and terminal outcomes from forked continuations on ScienceWorld and HotpotQA. We include an episode when every planned arm has a terminal summary and at least one priced continuation call, with no explicit arm error. Native failures and study limits with priced continuations remain in this ratio population; completed zero-call outcomes are reported separately below. This defines a comparison conditional on reaching the fork and completing the planned set of arms; it is not a benchmark-wide success rate. Arms run in fixed plan order. Six planned first-panel 8B episodes and seven observation-window episodes have no terminal summary and stop before Drop is observed. The reset procedure replays prefix actions but does not compare reset observations or state hashes with the base; exact environment-state equivalence is therefore not verified by these logs. Calls belonging to the shared prefix are excluded from continuation cost.

The principal ScienceWorld collections attempt five variations per task family. Their completed eligible panels are Qwen3-8B: 103 forks; Qwen3-4B: 126; Qwen3-30B-A3B: 119; Qwen3-32B: 52. They fork after five calls and stop at 30 calls including the prefix, with a 3,000-token generation cap and temperature zero. The 32B panel covers 12 families; the other three cover 30. Collections at forks three and eight and temperature 0.6 are separate controls. Retention is implemented in the raw completion prompt; it is not inferred from a model's default chat endpoint. In the Qwen renderer, each call produces a thinking block and a visible action, then the environment's returned text is serialized as a user message. The retention flags explicitly control all earlier thinking blocks. This is not a measurement of a provider's native within-turn tool-reasoning retention. Message roles and boundaries belong to the specified harness and rendering policy; an external result is not itself a reasoning block. The archived launchers identify \texttt{Qwen/Qwen3-4B} in BF16, \texttt{Qwen/Qwen3-30B-A3B-GPTQ-Int4} with eager execution, \texttt{Qwen/Qwen3-32B-AWQ} and \texttt{deepseek-ai/DeepSeek-R1-Distill-Llama-8B} in BF16. Surrounding launch scripts document \texttt{Qwen/Qwen3-8B} in BF16, but the original first-panel startup script and exact checkpoint revision are not archived. These are different model deployments, not a precision-matched architectural intervention.

The HotpotQA harness samples 200 questions without replacement from the distractor validation split, using seed 0 and the same initial sample for the four primary models. It preserves each question's original paragraph order. The agent opens paragraphs before submitting an answer. The primary transfer panels hide titles from the paragraph listing and instruct access by number; the action parser also accepts guessed titles. They fork after two calls and allow twelve calls in total. For 4B, 8B, MoE, and gpt-oss, completed populations are 122, 165, 88, and 195 respectively. Their context limits are 16,384 tokens for the three Qwen models and 32,768 for gpt-oss. Earlier answers terminate before the fork and are excluded. The harness awards 100 for exact match, otherwise a rounded token-F1 score; the archived comparison uses score 100. Only the answer is graded, without a supporting-fact criterion. Score 100 is full answer score, not a reconstructed exact-match indicator. The titles-visible 8B panel, which allows ten calls, and the completed second titles-hidden collection are reported separately.

The statistics use 4,000 bootstrap replicates. The original collection summaries use seed 20261011; the completed-window comparisons, joint cost/outcome tables, and reference-sensitivity analyses use seed 20261012. ScienceWorld resamples families, retaining all eligible variations and arms in each selected family. HotpotQA resamples question IDs. Main-body cost ratios use ratios of sums, corresponding to mean costs over each collected population; all numerators and denominators are resampled together. The full collection inventory additionally reports medians of within-episode ratios. Cross-environment comparisons have different eligibility filters and do not identify an isolated causal effect of environment.

The completed 8B truncation collection contributes 111 six-arm episodes; the second title-hidden 8B HotpotQA collection contributes 164 four-arm episodes. A further 171-fork HotpotQA collection crosses full history and the client window with Keep and Drop, alongside a second Keep execution. Each collection contributes its completed arm sets at the final archive cutoff. An incomplete 1.7B pilot contributes no complete arm set. The heavy-padding collection is retained as a failed intervention in the inventory.

The later five-arm sequential-task collection uses 32 matched chains, shuffled arm order and task-level resets. Its protocol and chain-bootstrap results are reported separately in Appendix~\ref{app:econ-sequential}.

\paragraph{Prompt and action contract.}
The ScienceWorld instruction asks for exactly one action on one line, no explanation outside thinking, and brief thinking before acting. Historical Qwen assistant messages contain a \texttt{<think>} block followed by the action; new external input is a user message. Drop leaves the thinking delimiters empty, while the delimiter ablation removes them. Generation starts after an open \texttt{<think>} tag and stops at the message terminator. The parser extracts the thought and the first non-note visible action line. Environment text is canonicalized by sorting comma-separated object lists and indented item lists before rendering; this stabilizes ordering but does not verify state equality. The note intervention appends the same instruction at each new call: write a visible \texttt{Note:} line of at most 40 words before the action. The parser retains at most 400 characters of that line. A note is therefore neither a guaranteed 40-word output nor a separate summarization call.

gpt-oss uses the Harmony format, whose official template drops analysis from completed earlier turns. Our Keep instead renders each earlier analysis message before its final message, and Drop leaves an empty analysis message, the analogue of Qwen's empty thinking block. Both arms are therefore custom cross-turn renderings: the gpt-oss comparisons do not separate retained reasoning content from the formatting of earlier turns, and in the 8B delimiter control, formatting alone moves the median cost ratio from 1.43 to 7.31.

\subsection{GPU execution and operation counting}
\label{app:econ-accounting}

\begin{figure}[!t]
\centering
\includegraphics[width=\columnwidth]{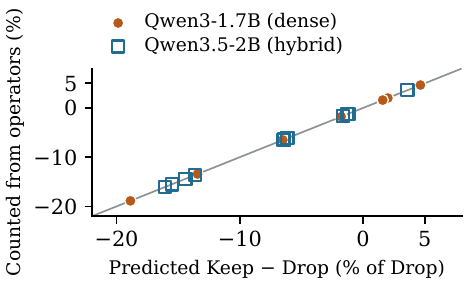}
\caption{\textbf{Predictions agree with GPU execution.} Predictions fixed before execution versus operation counts from instrumented vLLM: 27 dense/hybrid schedules, 648 requests. Counts cover matrix multiplications, the output head, causal attention pairs and recurrent-state updates. Both axes show the Keep-minus-Drop difference.}
\label{fig:econ-validation-parity}
\end{figure}

\begin{table}[t]\centering\small\setlength{\tabcolsep}{4pt}
\begin{tabular}{@{}lrrr@{}}\toprule
Model & Block & Requests & Max. error \\ \midrule
Qwen3-1.7B & 16 & 264 & 0 \\
Qwen3.5-2B & 544 & 384 & 0 \\
\bottomrule\end{tabular}
\caption{Predicted versus engine-recorded work. Block sizes are in tokens. The last column gives the largest absolute discrepancy in evaluated-token, attended-pair, or cached-token counts across all requests.}\label{tab:econ-validation}
\end{table}

The archived replay executes prescribed reasoning, answer, and observation token sequences under vLLM, BF16, on an RTX 3070. Qwen3-1.7B has 28 full-attention layers and a 16-token prefix block; Qwen3.5-2B has six full-attention and 18 recurrent layers and a 544-token block. Dense primary and replica sets contain 168 and 96 requests; hybrid sets contain 288 and 96. We recompute prefix matches from token IDs, apply the archived block-alignment rule, and compare the predictions with cached-token receipts and operations recorded during GPU execution.

Let $G_{\mathrm{out}}$ count returned output tokens. The final returned token has not yet been forwarded through the transformer, so these instrumented requests have $N=P+G_{\mathrm{out}}-1-R$ and $Q=\operatorname{tri}(P+G_{\mathrm{out}}-1)-\operatorname{tri}(R)$, where $\operatorname{tri}(x)=x(x+1)/2$. Every request has zero residual. Independently, each package froze predicted counts before execution. We multiply those counts by its matrix, output-head, and attention coefficients, adding its predicted recurrent-state subtotal, then compare with executed matrix shapes, causal attention pairs, and state-update operations. All 648 covered-operation totals match, across 11 dense and 16 hybrid schedules. Figure~\ref{fig:econ-validation-parity} reports this agreement for the covered operators. FLOPs are calculated from their recorded shapes and invocations; kernel overhead and memory traffic fall outside this count. Elapsed-time measurements appear separately in Appendix~\ref{app:econ-timing}.

The fork-study reduced prices instead use the producer's recorded $G$ convention and omit the head/state additions. Subtracting the unforwarded final output position while holding reconstructed cache counts fixed changes aggregate Drop/Keep ratios by at most 0.26\% across the four primary panels and three cache settings. No aggregate winner changes; one individual 4B full-reuse ranking changes. This isolates a boundary convention, not uncertainty about actual cache state.

The open-model prices take $c$ and $\beta$ from each archived run plan. They are reduced coefficients. In particular, the gpt-oss comparison does not add a complete sliding-window-layer operation account to its recorded full-attention coefficient. Its outcome contrast requires no FLOP assumption. Neither its cost ratio nor those of the other models are measurements of GPU time or money. The $w\in[1,5]$ sweep is a sensitivity range for generation pricing, not a fitted hardware measurement.

\subsection{Aligned mechanical replay}
\label{app:econ-replay}

With Keep as policy 1 and Drop as baseline, we compute $K=C(\mathrm{Keep};\xi_K)$, $D_K=C(\mathrm{Drop};\xi_K)$, and $D=C(\mathrm{Drop};\xi_D)$. Thus $\Delta M=K-D_K$, $\Delta B=D_K-D$, and $\Delta C=K-D$. The arithmetic identity is checked for every episode. To change reference, also compute $K_D=C(\mathrm{Keep};\xi_D)$. Then $\Delta M_D=K_D-D$ and $\Delta B_D=K-K_D$, with the same total $K-D$. Table~\ref{tab:econ-reference-sensitivity} reports both allocations. For example, the 8B full-reuse mechanical term changes from $-1.5\%$ to $14.0\%$ of Keep cost. The component shares depend on the chosen continuation reference; their sum is invariant.

Drop removes all earlier reasoning, including the shared-prefix thoughts, exactly as in the live runner. Before pricing a replay, we reconstruct each actual Keep and Drop prompt and verify its token count and SHA-1 hash. Across the four included panels, 13,521 actual prompts pass both checks. One MoE fork has different task text across resets; its available base text fails to reproduce 18 actual-arm prompt hashes despite matching token counts. It remains in the 119-fork live comparison and is excluded only from replay, leaving 118 MoE replay pairs.

Concretely, let $s_t$ be the tokenized rendered prompt and $z_{t-1}$ the preceding cache-eligible sequence. Write $\operatorname{lcp}$ for their common-prefix length. Each reconstructed request uses
\[
 P_t=|s_t|,\qquad R_t=16\left\lfloor\frac{\operatorname{lcp}(s_t,z_{t-1})}{16}\right\rfloor.
\]
For prompt-only reuse set $z_t=s_t$. For generated-state reuse, concatenate the prompt and canonical completion and retain at most its first $P_t+G_t$ token positions. Apply Equation~\ref{eq:econ-app-call} at each call and sum. The three replays start from the same shared-prefix state.

The call archive retains parsed reasoning and actions, rather than every byte of the original wire completion. The generated-state-reuse replay therefore caches the canonical completion built from those parsed fields, capped at the preceding call's charged $P+G$ positions. This cap also applies to the shared prefix: a reconstructed closing delimiter cannot create reusable positions that were never charged. The prompt-only reconstruction instead caches the preceding rendered prompt, with no reconstruction of completion text. Both use contiguous prefix matches and 16-token alignment. Within each reconstruction, all three costs $K,D_K,D$ use the same rules. These are specified counterfactual cache models, not measurements of the server's KV state. The logged native cached-token field maps missing values to zero, so a recorded zero does not establish a zero cache hit. The prices use reconstructed counts, not that ambiguous receipt field.

The repricing map uses the logged reconstruction on each arm, caps its reusable prefix at the previous prompt for prompt-only reuse, and sets it to zero for no reuse. It sweeps $P^*\in[2{,}500,250{,}000]$ and $w\in[1,5]$. Changing this price surface keeps both continuations fixed. The aligned replay performs a different operation: it changes the rendered history on the chosen reference continuation. These two analyses answer different questions and their cost estimates are not mixed.

\begin{table}[htbp]\centering\small
\setlength{\tabcolsep}{4pt}
\begin{tabular}{@{}llrrrrr@{}}\toprule
Model & Reuse & $\Delta M_K$ & $-\Delta B_K$ & $\Delta M_D$ & $-\Delta B_D$ & $\Delta C$\\\midrule
8B & Prompt + generation & -1.5 & 73.3 & 14.0 & 88.8 & -74.8\\
8B & Prompt only & 40.5 & 42.8 & 95.8 & 98.1 & -2.3\\
4B & Prompt + generation & 38.6 & 56.7 & 95.8 & 113.9 & -18.1\\
4B & Prompt only & 67.3 & 30.1 & 154.6 & 117.4 & 37.2\\
30B-A3B & Prompt + generation & 27.4 & 100.8 & 144.8 & 218.3 & -73.5\\
30B-A3B & Prompt only & 60.0 & 55.5 & 235.3 & 230.8 & 4.5\\
32B & Prompt + generation & 4.6 & 98.8 & 35.3 & 129.5 & -94.2\\
32B & Prompt only & 48.6 & 53.2 & 129.6 & 134.3 & -4.7\\
\bottomrule\end{tabular}
\caption{Changing the replay reference changes the allocation. Subscripts identify the Keep or Drop continuation used to price the mechanical term. Entries are aggregate percentages of Keep cost; $\Delta C=\Delta M-(-\Delta B)$ under either reference. The total and repayment side are invariant by construction. The panels contain 103, 126, 118, and 52 forks. Totals come from the aligned replay (Appendix~\ref{app:econ-replay}), which prices Keep 1.8\% below the logged reconstruction; for 8B with generated-state reuse its $\Delta C$ implies Drop/Keep of 1.748, against 1.738 logged (Table~\ref{tab:econ-estimand-comparison}).}\label{tab:econ-reference-sensitivity}
\end{table}

\subsection{Client-window replay}
\label{app:econ-window-replay}

Within each reasoning-retention setting, write W for the client window and F for full history. Reconstruct four costs from the two recorded continuations:
\[
 C(W;\xi_W),\ C(F;\xi_W),\ C(F;\xi_F),\ C(W;\xi_F).
\]
Using $\xi_W$ as reference gives $\Delta M_W=C(W;\xi_W)-C(F;\xi_W)$ and $\Delta B_W=C(F;\xi_W)-C(F;\xi_F)$. Using $\xi_F$ gives $\Delta M_F=C(W;\xi_F)-C(F;\xi_F)$ and $\Delta B_F=C(W;\xi_W)-C(W;\xi_F)$. Both sum to the observed-continuation cost difference under the specified replay. Windowing changes the rendered environment-result history; each retained reasoning block follows the same Keep or Drop rule in all four cells. Generation lengths, actions, and returned environment text stay fixed within each replay reference.

All 6,594 actual-arm ScienceWorld prompts and 2,702 HotpotQA prompts match their archived hashes and token counts; no eligible grid fork is excluded from replay. Each four-cell comparison starts from the same shared-prefix cache state and uses the prefix matching and completion boundary above. Prompt agreement verifies rendering, not environment-reset equivalence. Table~\ref{tab:econ-window-repayment} gives the full allocation under both cache reconstructions.

\begin{table}[htbp]\centering\small
\setlength{\tabcolsep}{4pt}
\begin{tabular}{@{}lllrrrrr@{}}\toprule
Task & Reuse & Reasoning & $\Delta M_W$ & $-\Delta B_W$ & $\Delta M_F$ & $-\Delta B_F$ & $\Delta C$\\\midrule
ScienceWorld & Prompt + generation & Keep & 338.4 & -14.6 & 301.9 & -51.0 & 352.9\\
ScienceWorld & Prompt + generation & Drop & 35.7 & -9.7 & 38.6 & -6.7 & 45.3\\
ScienceWorld & Prompt only & Keep & 141.1 & -13.9 & 126.3 & -28.7 & 155.0\\
ScienceWorld & Prompt only & Drop & 34.1 & -9.4 & 36.9 & -6.6 & 43.5\\
HotpotQA & Prompt + generation & Keep & 146.2 & 0.3 & 140.3 & -5.6 & 145.9\\
HotpotQA & Prompt + generation & Drop & 36.2 & 5.1 & 47.4 & 16.4 & 31.0\\
HotpotQA & Prompt only & Keep & 66.9 & -0.3 & 63.4 & -3.7 & 67.2\\
HotpotQA & Prompt only & Drop & 35.2 & 5.2 & 46.2 & 16.2 & 30.0\\
\bottomrule\end{tabular}
\caption{Client-window repayment under two continuation references. W uses the window continuation and F the full-history continuation. Every entry is a percentage of aggregate full-history cost with the same reasoning-retention rule. Policy 1 is the client window and the baseline the full history; positive $-\Delta B$ is a continuation saving, and $\Delta C=\Delta M-(-\Delta B)$. The mechanical term exceeds the magnitude of the behavioral term in every displayed configuration and under both references. The 114 ScienceWorld and 171 HotpotQA forks reproduce all 9,296 actual-arm prompt hashes and token counts.}\label{tab:econ-window-repayment}
\end{table}

\subsection{Collection flow and stopping rules}

The entries of Table~\ref{tab:econ-flow} distinguish planning, reaching the fork, completed arm sets, and the stricter cost-ratio population. ``No terminal record'' denotes a missing episode summary at this archive snapshot; it is not scored as a failure or assumed missing at random.

\small

\begin{table*}[tp]\centering\small
\begin{tabular}{@{}p{.38\textwidth}rrrrr@{}}\toprule
Collection & Planned & Pre-fork & Terminal & Cost pairs & Missing \\ \midrule
8B ScienceWorld, sequential repeat & 40 & 6 & 34 & 34 & 0 \\
4B ScienceWorld, thought truncation & 149 & 28 & 121 & 116 & 0 \\
8B ScienceWorld, thought truncation & 149 & 38 & 111 & 111 & 0 \\
8B ScienceWorld, fork at call 3 & 149 & 15 & 134 & 134 & 0 \\
8B ScienceWorld, fork at call 8 & 149 & 59 & 90 & 90 & 0 \\
gpt-oss ScienceWorld & 100 & 12 & 88 & 88 & 0 \\
30B-A3B ScienceWorld & 149 & 30 & 119 & 119 & 0 \\
8B ScienceWorld, delimiters & 149 & 42 & 107 & 107 & 0 \\
8B ScienceWorld, 12K padding & 60 & 14 & 46 & 46 & 0 \\
8B ScienceWorld, 16K padding & 50 & 6 & 44 & 43 & 0 \\
8B ScienceWorld, primary & 149 & 40 & 103 & 103 & 6 \\
1.7B ScienceWorld pilot & 149 & 7 & 0 & 0 & 142 \\
32B ScienceWorld & 60 & 8 & 52 & 52 & 0 \\
4B ScienceWorld, primary & 149 & 23 & 126 & 126 & 0 \\
R1-distill ScienceWorld & 100 & 5 & 95 & 95 & 0 \\
8B ScienceWorld, memory note & 100 & 27 & 73 & 73 & 0 \\
8B ScienceWorld, memory note repeat & 100 & 19 & 81 & 81 & 0 \\
8B ScienceWorld, temperature 0.6 & 149 & 32 & 117 & 117 & 0 \\
8B ScienceWorld, input window & 149 & 28 & 114 & 114 & 7 \\
gpt-oss HotpotQA, hidden titles & 200 & 5 & 195 & 195 & 0 \\
30B-A3B HotpotQA, hidden titles & 200 & 112 & 88 & 88 & 0 \\
4B HotpotQA, hidden titles & 200 & 78 & 122 & 122 & 0 \\
8B HotpotQA, visible titles & 200 & 107 & 92 & 92 & 1 \\
8B HotpotQA, hidden titles & 200 & 35 & 165 & 165 & 0 \\
8B HotpotQA, hidden titles repeat & 200 & 36 & 164 & 164 & 0 \\
8B HotpotQA, input window & 200 & 29 & 171 & 171 & 0 \\
R1-distill HotpotQA, hidden titles & 200 & 144 & 56 & 56 & 0 \\
\bottomrule\end{tabular}
\caption{Collection flow. Pre-fork ends have no continuation comparison. Terminal sets have every arm completed; cost pairs additionally require a priced call in every arm. Missing ends are incomplete episode records.}\label{tab:econ-flow}
\end{table*}

\normalsize

The client limits generation by the remaining context budget, reserving eight tokens; it stops an arm as context-full when fewer than 64 generation tokens remain. Such an arm retains its last native score. This stopping decision is separate from the 3,000-token output cap and total-call limit. A context cap is an intervention-relevant outcome, not an infrastructure exception to delete from success accounting. Plans with no context-limit field do not establish the original driver's effective limit; launch scripts establish engine limits only for their documented deployments.

\subsection{Mechanical retention share under two cache rules}
\label{app:econ-retention-share}

\begin{figure}[!t]
\centering
\includegraphics[width=\columnwidth]{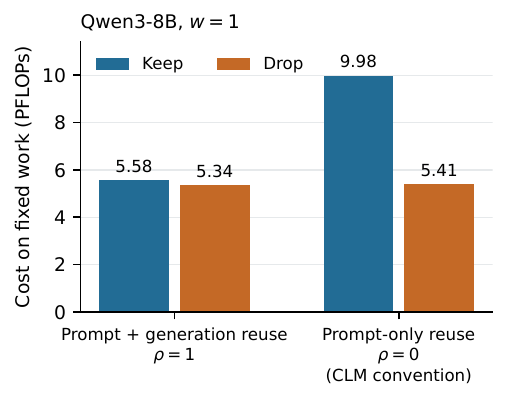}
\caption{\textbf{Re-reading history changes its cost.} The same 134 Qwen3-8B ScienceWorld Keep continuations, from the fork-three collection rather than the primary 103-fork panel, replayed with earlier reasoning retained or removed at $w=1$. Each bar sums the cost of the entire panel. Prompt-only reuse ($\rho=0$), the standard-serving convention in \citet[Appendix~C]{shao2026contextlanguagemodels}, reprocesses returning output. Retention adds 45.8\% of Keep's cost under this rule, versus 4.2\% with generated-state reuse.}
\label{fig:econ-cache-cost}
\end{figure}

The fork-three collection contains 134 Qwen3-8B ScienceWorld episodes. We replay Keep's continuation under Keep and Drop; Drop removes both pre-fork and post-fork reasoning, matching the live policy. The same tokenizer and 16-token prefix alignment are used in both configurations. Full reuse caches canonical parsed completions up to the preceding charged $P+G$ length; prompt-only reuse caches the previous rendered prompt. Every actual-arm prompt matches its archived token count and hash: 4,104 checks.

For each cache configuration, the aggregate share is
\[
 \frac{\sum_i\{C(\mathrm{Keep};\xi_{K,i})-C(\mathrm{Drop};\xi_{K,i})\}}
      {\sum_i C(\mathrm{Keep};\xi_{K,i})}.
\]
It is 4.2\% [1.0, 7.0] with generated-state reuse and 45.8\% [42.9, 48.4] with prompt-only reuse. The corresponding medians of episode shares are $-2.1\%$ and 40.7\%; these are different estimands. The primary 103-pair fork-five panel gives aggregate shares $-1.5\%$ and 40.5\%, with 2,992 actual-arm prompt checks. Negative values denote a mechanical saving on the chosen reference. These allocations price specified cache replays; they are not measured monetary bills or reference-independent causal shares of the live policy difference.
\section{Uncertainty, controls, and full outcomes}
\label{app:econ-controls}

The complete collection table (Table~\ref{tab:econ-inventory}) preserves controls and unsuccessful interventions. Keep2 is a second execution of Keep from the same fork. It measures run-to-run variability under that setup. The Rule arm uses an approximate mechanical threshold. For a newly returned response, let $r$ be the tokenized parsed thought and $a=\max(1,G-r)$; this residual includes formatting positions. At prompt length $P$, it keeps the thought when
\[
 (T_{\max}-t)\overline N_{<t+1}
 < \frac{a(P^*+P+a/2)}{\max(1,r)}.
\]
Here $T_{\max}$ is the episode's call limit, $t$ is the number of calls before the current response and $\overline N_{<t+1}$ is the mean evaluated-token count over those calls, including the shared prefix. The horizon multiplier therefore includes one extra nominal call; the threshold omits the discrete $+1/2$ term. This is the implemented heuristic, not an exact application of the fixed-continuation rule, and it contains no estimate of the behavioral term. It assumes generated-state reuse, which the vLLM server used for these runs provides.

The note experiments ask for a visible memory note before the action. Note + Keep retains reasoning and the note; Note + Drop drops reasoning and retains the note. Both alter the prompt and subsequent generation. The two collections use the same 40-word note instruction; the second is a same-settings replicate. The separate chained-task study compares automatic summarization with model-directed editing (Appendix~\ref{sec:econ-client-factorial}).

The dose collections retain at most the first 150, 40, or 10 tokens of each thought, or replace the thought with a token-length-targeted sequence assembled from another run's trace pool. The swap code does not enforce a different source episode; it concatenates pool entries as needed and decodes a truncated token sequence. Truncation changes both content and length. Later outputs also reflect different actions and states; they do not identify a content-independent effect of reasoning length. The swap/Keep median paired cost ratios are 2.55 for 4B and 5.15 for 8B. The shorter-prefix arms have different cost and full-score outcomes across the models. The model-specific outcomes motivate evaluating both the amount and content of retained reasoning.

The R1-distill ScienceWorld collection has zero full-score continuations in all three arms across 95 forks. With empty reasoning blocks, median reasoning falls from 380 tokens per call under Keep to one, accompanied by a median Drop/Keep cost ratio of 0.21. This establishes no successful policy improvement. The longer-context padding collections also change input content; repetitive-padding degeneration is reported as such, not as a clean architectural crossover.

The cost-ratio population requires a priced continuation in every arm. Consequently, it excludes five additional completed six-arm 4B dose episodes whose Keep and Swap arms hit the context cap before a first continuation call, and one such heavy-padding episode. Drop reaches full score in one of the five excluded dose episodes. These exclusions affect the control panels, not the four principal replay panels; the reported dose ratios describe the 116 positive-cost complete-arm episodes, rather than all 121 completed six-arm episodes.

\subsection{Full outcomes for zero-call episodes}

Table~\ref{tab:econ-terminal-outcomes} restores all completed native outcomes for the two collections with zero-call exclusions. The left outcome column includes those arms; the right uses exactly the denominator of the published cost-ratio inventory. No undefined ratio is replaced by zero or infinity.

\begin{table}[ht]\centering\small\setlength{\tabcolsep}{4pt}
\begin{tabular}{@{}p{.35\textwidth}lrrr@{}}\toprule
Collection & Policy & All terminal & Ratio panel & Zero calls \\ \midrule
4B ScienceWorld, thought truncation & Keep & 11/121 & 11/116 & 5 \\
 & First 150 & 11/121 & 11/116 & 0 \\
 & First 40 & 16/121 & 16/116 & 0 \\
 & First 10 & 9/121 & 9/116 & 0 \\
 & Swap & 7/121 & 7/116 & 5 \\
 & Drop & 16/121 & 15/116 & 0 \\
8B ScienceWorld, 16K padding & Keep & 5/44 & 5/43 & 1 \\
 & Keep repeat & 7/44 & 7/43 & 1 \\
 & Drop & 3/44 & 3/43 & 0 \\
 & Rule & 5/44 & 5/43 & 1 \\
\bottomrule\end{tabular}
\caption{Full-score outcomes with and without the positive-cost inclusion rule. Zero-call counts are arm-specific.}\label{tab:econ-terminal-outcomes}
\end{table}

The four principal repayment panels have no additional complete sets excluded by this zero-call rule.

\subsection{Typical ratios and mean cost}

For paired costs $(K_i,D_i)$, the median of $D_i/K_i$ describes a typical within-pair ratio; $\sum_iD_i/\sum_iK_i$ compares average cost over the same panel. The latter weights individual ratios by $K_i$. They need not lie on the same side of one: in the 73-pair memory-note collection, Drop/Keep is 1.652 by median ratio and 0.947 by total expenditure. The values in Table~\ref{tab:econ-estimand-comparison} are recomputed from the recorded per-call account, with the same $w=1$ price in both arms.

\begin{table}[ht]\centering\small\setlength{\tabcolsep}{4pt}
\begin{tabular}{@{}p{.38\textwidth}lrrr@{}}\toprule
Collection & Reuse & Pairs & Median D/K & Total D/K \\ \midrule
8B ScienceWorld, primary & full & 103 & 1.871 & 1.738 \\
8B ScienceWorld, primary & prompt & 103 & 1.036 & 1.023 \\
4B ScienceWorld, primary & full & 126 & 1.704 & 1.181 \\
4B ScienceWorld, primary & prompt & 126 & 0.959 & 0.628 \\
30B-A3B ScienceWorld & full & 119 & 1.867 & 1.729 \\
30B-A3B ScienceWorld & prompt & 119 & 1.093 & 0.954 \\
8B ScienceWorld, memory note & full & 73 & 1.652 & 0.947 \\
8B ScienceWorld, memory note & prompt & 73 & 0.787 & 0.530 \\
8B HotpotQA, hidden titles & full & 165 & 0.979 & 0.812 \\
8B HotpotQA, hidden titles & prompt & 165 & 0.652 & 0.517 \\
\bottomrule\end{tabular}
\caption{Median paired ratios and ratios of total costs over identical completed panels. Reuse includes generation (full) or only the prompt.}\label{tab:econ-estimand-comparison}
\end{table}

Figures~\ref{fig:econ-repricing} and~\ref{fig:econ-repayment} use sums to compare mean cost over their collected populations. The episode medians above describe typical within-pair ratios; they do not determine the minimum-expenditure choice.

\begin{table}[htbp]\centering\small
\setlength{\tabcolsep}{4pt}
\begin{tabular}{@{}lll@{}}\toprule
Reused prefix & Median D / K & Total D / K\\\midrule
Prompt + generation & 1.87 [1.38, 2.29] & 1.74 [1.46, 2.08]\\
Prompt only & 1.04 [0.79, 1.40] & 1.02 [0.86, 1.23]\\
None & 0.40 [0.30, 0.49] & 0.34 [0.28, 0.42]\\
\bottomrule\end{tabular}
\caption{Repricing the same 103 ScienceWorld 8B forks at $w=1$. Both estimands retain paired family-bootstrap uncertainty. The prompt-only intervals span equal cost; the full-reuse and no-reuse intervals lie on opposite sides of one. Intervals use bootstrap seed 20261012.}\label{tab:econ-repricing-uncertainty}
\end{table}
\begin{table}[htbp]\centering\small
\setlength{\tabcolsep}{4pt}
\begin{tabular}{@{}llrrll@{}}\toprule
Task & Model & Forks & Full-score K / D & Total cost D / K & Full-score D $-$ K (pp)\\\midrule
SW & 8B & 103 & 16 / 12 & 1.74 [1.46, 2.08] & -3.9 [-10.2, 2.0]\\
SW & 4B & 126 & 17 / 18 & 1.18 [0.93, 1.49] & 0.8 [-8.8, 9.0]\\
SW & 30B-A3B & 119 & 19 / 18 & 1.73 [1.43, 2.09] & -0.8 [-6.6, 5.4]\\
SW & 32B & 52 & 7 / 7 & 1.94 [1.65, 2.26] & 0.0 [0.0, 0.0]\\
SW & gpt-oss-20b & 88 & 3 / 20 & 1.20 [0.96, 1.49] & 19.3 [10.7, 29.3]\\
HP & 8B & 165 & 64 / 42 & 0.81 [0.64, 1.00] & -13.3 [-20.6, -6.1]\\
HP repeat & 8B & 164 & 66 / 48 & 0.87 [0.75, 1.00] & -11.0 [-18.9, -3.7]\\
HP & 4B & 122 & 47 / 30 & 0.70 [0.48, 1.04] & -13.9 [-22.1, -5.7]\\
HP & 30B-A3B & 88 & 39 / 19 & 0.92 [0.80, 1.06] & -22.7 [-33.0, -11.4]\\
HP & gpt-oss-20b & 195 & 41 / 23 & 0.92 [0.78, 1.09] & -9.2 [-14.9, -4.1]\\
\bottomrule\end{tabular}
\caption{Cost and outcome on the same eligible forks. SW: ScienceWorld; HP: title-hidden HotpotQA. Models are Qwen3 unless named otherwise. Cost ratios divide total Drop cost by total Keep cost at the recorded reduced price; bracketed values are paired 95\% cluster-bootstrap intervals. A zero interval for the 32B outcome contrast records identical paired outcomes in this panel, not population equivalence.}\label{tab:econ-joint-outcomes}
\end{table}

\subsection{The repeated HotpotQA collection}

The two title-hidden 8B HotpotQA collections draw on the same 200-question sample and share 153 eligible questions. Table~\ref{tab:econ-joint-outcomes} reports each full panel: $-13.3$ points on 165 questions and $-11.0$ on 164. Restricted to the 153 shared questions, Keep/Drop full-score counts are 61/41 in the first and 64/44 in the repeat, a paired Drop-minus-Keep difference of $-13.1$ points in each, with question-bootstrap intervals [$-20.9$, $-5.2$] and [$-20.9$, $-5.9$]. The repetition reproduces the direction within this harness and sample; it is not an independent question sample.

\subsection{Complete cost-ratio inventory}

Table~\ref{tab:econ-inventory} reports each positive-cost complete-arm collection. Costs are median paired ratios to Keep with 95\% cluster-bootstrap intervals. Full-score counts use that same denominator; Table~\ref{tab:econ-terminal-outcomes} supplies the wider outcome population wherever zero-call exclusions matter. Collections are identified by model, benchmark, and intervention. Qwen sizes identify the Qwen3 family; R1-distill denotes DeepSeek-R1-Distill-Llama-8B and gpt-oss denotes gpt-oss-20b.

\small

\begin{table*}[tp]\centering\small
\begin{tabular}{@{}p{.30\textwidth}lrrrr@{}}\toprule
Collection & Policy & Forks & Full-score & Cost / Keep & 95\% interval \\ \midrule
8B ScienceWorld, sequential repeat & Keep & 34 & 2 & 1.00 & [1.00, 1.00] \\
 & Keep repeat & 34 & 2 & 1.00 & [0.98, 1.14] \\
 & Drop & 34 & 2 & 1.27 & [0.84, 2.96] \\
 & Rule & 34 & 3 & 0.98 & [0.84, 1.37] \\
\midrule
4B ScienceWorld, thought truncation & Keep & 116 & 11 & 1.00 & [1.00, 1.00] \\
 & First 150 & 116 & 11 & 0.99 & [0.81, 1.10] \\
 & First 40 & 116 & 16 & 0.81 & [0.65, 0.98] \\
 & First 10 & 116 & 9 & 0.77 & [0.59, 0.90] \\
 & Swap & 116 & 7 & 2.55 & [2.15, 3.35] \\
 & Drop & 116 & 15 & 1.92 & [1.57, 2.26] \\
\midrule
8B ScienceWorld, thought truncation & Keep & 111 & 14 & 1.00 & [1.00, 1.00] \\
 & First 150 & 111 & 20 & 1.41 & [1.14, 1.83] \\
 & First 40 & 111 & 20 & 1.15 & [0.90, 1.42] \\
 & First 10 & 111 & 20 & 0.84 & [0.67, 1.00] \\
 & Swap & 111 & 6 & 5.15 & [4.33, 7.05] \\
 & Drop & 111 & 14 & 1.59 & [1.25, 1.89] \\
\midrule
8B ScienceWorld, fork at call 3 & Keep & 134 & 25 & 1.00 & [1.00, 1.00] \\
 & Keep repeat & 134 & 22 & 1.00 & [0.92, 1.00] \\
 & Drop & 134 & 28 & 1.44 & [1.15, 1.66] \\
 & Rule & 134 & 24 & 1.14 & [1.03, 1.52] \\
\midrule
8B ScienceWorld, fork at call 8 & Keep & 90 & 10 & 1.00 & [1.00, 1.00] \\
 & Keep repeat & 90 & 7 & 1.00 & [0.98, 1.00] \\
 & Drop & 90 & 10 & 1.54 & [0.93, 2.06] \\
 & Rule & 90 & 10 & 1.00 & [1.00, 1.04] \\
\midrule
gpt-oss ScienceWorld & Keep & 88 & 3 & 1.00 & [1.00, 1.00] \\
 & Drop & 88 & 20 & 1.31 & [0.96, 1.75] \\
 & Drop, no tags & 88 & 15 & 1.59 & [1.25, 2.06] \\
\midrule
30B-A3B ScienceWorld & Keep & 119 & 19 & 1.00 & [1.00, 1.00] \\
 & Keep repeat & 119 & 18 & 0.98 & [0.94, 1.01] \\
 & Drop & 119 & 18 & 1.87 & [1.58, 2.27] \\
 & Rule & 119 & 20 & 1.26 & [1.13, 1.44] \\
\midrule
8B ScienceWorld, delimiters & Keep & 107 & 13 & 1.00 & [1.00, 1.00] \\
 & Drop & 107 & 15 & 1.43 & [1.23, 1.76] \\
 & Drop, no tags & 107 & 21 & 7.31 & [5.79, 8.75] \\
\midrule
8B ScienceWorld, 12K padding & Keep & 46 & 2 & 1.00 & [1.00, 1.00] \\
 & Keep repeat & 46 & 2 & 1.04 & [0.92, 1.26] \\
 & Drop & 46 & 2 & 0.92 & [0.73, 1.12] \\
 & Rule & 46 & 2 & 1.11 & [0.80, 1.33] \\
\midrule
8B ScienceWorld, 16K padding & Keep & 43 & 5 & 1.00 & [1.00, 1.00] \\
 & Keep repeat & 43 & 7 & 1.00 & [1.00, 1.00] \\
 & Drop & 43 & 3 & 6.57 & [2.75, 21.33] \\
 & Rule & 43 & 5 & 3.98 & [1.05, 5.05] \\
\midrule
8B ScienceWorld, primary & Keep & 103 & 16 & 1.00 & [1.00, 1.00] \\
 & Keep repeat & 103 & 19 & 1.00 & [0.99, 1.00] \\
 & Drop & 103 & 12 & 1.87 & [1.41, 2.29] \\
 & Rule & 103 & 13 & 1.10 & [0.91, 1.32] \\
\bottomrule\end{tabular}
\caption{Full cost-ratio inventory. Cost / Keep is the median within-pair ratio at $w=1$ under the logged generated-state-reuse reconstruction; intervals resample families or questions with bootstrap seed 20261011, so they can differ in the second decimal from Tables~\ref{tab:econ-repricing-uncertainty} and~\ref{tab:econ-joint-outcomes}, which use seed 20261012.}\label{tab:econ-inventory}
\end{table*}

\begin{table*}[tp]\centering\small
\begin{tabular}{@{}p{.30\textwidth}lrrrr@{}}\toprule
Collection & Policy & Forks & Full-score & Cost / Keep & 95\% interval \\ \midrule
32B ScienceWorld & Keep & 52 & 7 & 1.00 & [1.00, 1.00] \\
 & Keep repeat & 52 & 4 & 1.00 & [1.00, 1.00] \\
 & Drop & 52 & 7 & 2.24 & [1.77, 2.70] \\
 & Rule & 52 & 6 & 1.26 & [0.99, 1.55] \\
\midrule
4B ScienceWorld, primary & Keep & 126 & 17 & 1.00 & [1.00, 1.00] \\
 & Keep repeat & 126 & 19 & 1.00 & [1.00, 1.00] \\
 & Drop & 126 & 18 & 1.70 & [1.36, 2.09] \\
 & Rule & 126 & 16 & 1.12 & [0.97, 1.26] \\
\midrule
R1-distill ScienceWorld & Keep & 95 & 0 & 1.00 & [1.00, 1.00] \\
 & Drop & 95 & 0 & 0.21 & [0.19, 0.26] \\
 & Drop, no tags & 95 & 0 & 0.77 & [0.68, 0.86] \\
\midrule
8B ScienceWorld, memory note & Keep & 73 & 15 & 1.00 & [1.00, 1.00] \\
 & Drop & 73 & 8 & 1.65 & [0.93, 1.82] \\
 & Note + Keep & 73 & 13 & 4.74 & [3.07, 5.66] \\
 & Note + Drop & 73 & 11 & 3.26 & [2.13, 5.22] \\
\midrule
8B ScienceWorld, memory note repeat & Keep & 81 & 16 & 1.00 & [1.00, 1.00] \\
 & Drop & 81 & 11 & 1.61 & [1.34, 2.40] \\
 & Note + Keep & 81 & 9 & 5.44 & [3.63, 7.00] \\
 & Note + Drop & 81 & 14 & 3.50 & [2.91, 4.94] \\
\midrule
8B ScienceWorld, temperature 0.6 & Keep & 117 & 20 & 1.00 & [1.00, 1.00] \\
 & Keep repeat & 117 & 18 & 1.00 & [0.99, 1.00] \\
 & Drop & 117 & 21 & 1.52 & [1.27, 1.79] \\
 & Rule & 117 & 17 & 1.11 & [1.00, 1.29] \\
\midrule
8B ScienceWorld, input window & Keep & 114 & 19 & 1.00 & [1.00, 1.00] \\
 & Drop & 114 & 20 & 1.56 & [1.19, 1.78] \\
 & Window + Keep & 114 & 13 & 4.28 & [3.64, 4.85] \\
 & Window + Drop & 114 & 18 & 2.06 & [1.80, 2.47] \\
\midrule
gpt-oss HotpotQA, hidden titles & Keep & 195 & 41 & 1.00 & [1.00, 1.00] \\
 & Keep repeat & 195 & 40 & 0.99 & [0.89, 1.00] \\
 & Drop & 195 & 23 & 0.97 & [0.82, 1.32] \\
 & Rule & 195 & 33 & 1.00 & [0.97, 1.03] \\
\midrule
30B-A3B HotpotQA, hidden titles & Keep & 88 & 39 & 1.00 & [1.00, 1.00] \\
 & Keep repeat & 88 & 39 & 1.02 & [1.00, 1.08] \\
 & Drop & 88 & 19 & 1.04 & [0.94, 1.17] \\
 & Rule & 88 & 35 & 1.08 & [1.02, 1.14] \\
\midrule
4B HotpotQA, hidden titles & Keep & 122 & 47 & 1.00 & [1.00, 1.00] \\
 & Keep repeat & 122 & 42 & 1.00 & [1.00, 1.01] \\
 & Drop & 122 & 30 & 0.95 & [0.74, 1.14] \\
 & Rule & 122 & 37 & 1.04 & [1.00, 1.07] \\
\midrule
8B HotpotQA, visible titles & Keep & 92 & 53 & 1.00 & [1.00, 1.00] \\
 & Keep repeat & 92 & 54 & 1.00 & [1.00, 1.00] \\
 & Drop & 92 & 47 & 1.20 & [0.98, 1.47] \\
 & Rule & 92 & 53 & 1.00 & [1.00, 1.00] \\
\midrule
8B HotpotQA, hidden titles & Keep & 165 & 64 & 1.00 & [1.00, 1.00] \\
 & Keep repeat & 165 & 62 & 1.00 & [1.00, 1.00] \\
 & Drop & 165 & 42 & 0.98 & [0.82, 1.11] \\
 & Rule & 165 & 57 & 1.00 & [1.00, 1.05] \\
\bottomrule\end{tabular}
\addtocounter{table}{-1}\caption{Full cost-ratio inventory (continued).}
\end{table*}

\begin{table*}[tp]\centering\small
\begin{tabular}{@{}p{.30\textwidth}lrrrr@{}}\toprule
Collection & Policy & Forks & Full-score & Cost / Keep & 95\% interval \\ \midrule
8B HotpotQA, hidden titles repeat & Keep & 164 & 66 & 1.00 & [1.00, 1.00] \\
 & Keep repeat & 164 & 62 & 1.00 & [0.99, 1.00] \\
 & Drop & 164 & 48 & 1.00 & [0.92, 1.18] \\
 & Rule & 164 & 56 & 1.04 & [1.00, 1.09] \\
\midrule
8B HotpotQA, input window & Keep & 171 & 57 & 1.00 & [1.00, 1.00] \\
 & Keep repeat & 171 & 63 & 1.00 & [1.00, 1.00] \\
 & Drop & 171 & 44 & 1.02 & [0.92, 1.15] \\
 & Window + Keep & 171 & 62 & 2.12 & [1.96, 2.39] \\
 & Window + Drop & 171 & 45 & 1.21 & [1.12, 1.37] \\
\midrule
R1-distill HotpotQA, hidden titles & Keep & 56 & 1 & 1.00 & [1.00, 1.00] \\
 & Keep repeat & 56 & 3 & 1.00 & [1.00, 1.00] \\
 & Drop & 56 & 2 & 0.34 & [0.15, 0.75] \\
 & Rule & 56 & 3 & 0.85 & [0.70, 0.91] \\
\bottomrule\end{tabular}
\addtocounter{table}{-1}\caption{Full cost-ratio inventory (continued).}
\end{table*}

\normalsize

\subsection{Mechanical change against continuation saving}

\begin{figure}[htbp]\centering
\includegraphics[width=.97\textwidth]{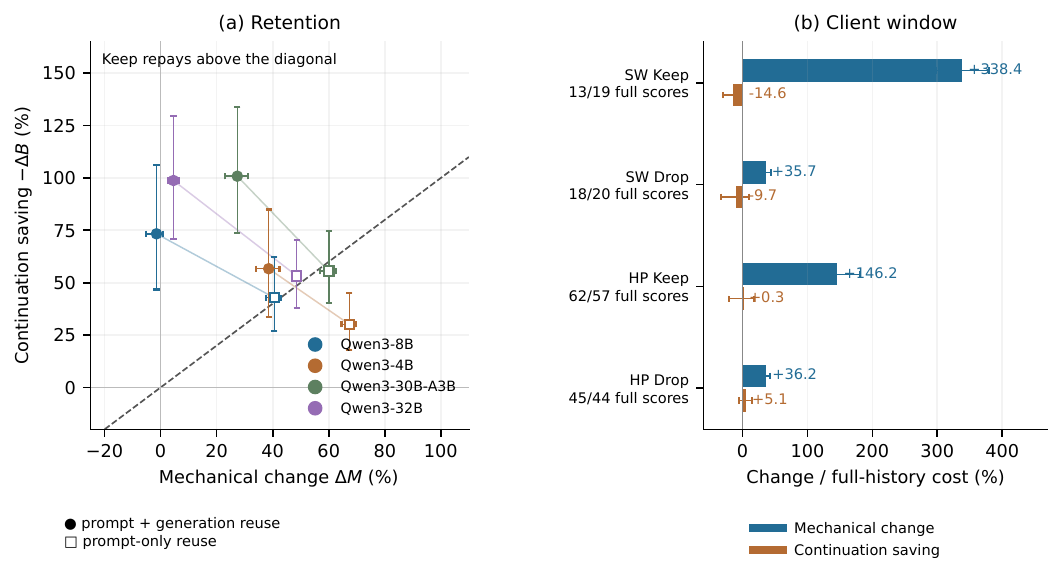}
\caption{Mechanical change and continuation saving under the declared cache reconstructions, normalized by the reference population's cost. The retention panels contain 103/126/118/52 pairs for 8B/4B/MoE/32B; the window panels contain 114 ScienceWorld and 171 HotpotQA forks. Positive continuation savings offset mechanical expense. Intervals resample task families or questions.}\label{fig:econ-repayment}
\end{figure}

Figure~\ref{fig:econ-repayment} plots the two sides of Equation~\ref{eq:econ-repay} for every retention and window panel, normalized by the reference population's cost; it is the graphical form of Tables~\ref{tab:econ-reference-sensitivity} and~\ref{tab:econ-window-repayment}.

\section{Hosted runs, trace evidence and provider rules}
\subsection{Original hosted collections}
\label{app:econ-hosted}

The hosted Qwen3.7-Plus ScienceWorld collection has 159 fork pairs from 58 tasks in 30 families, including repetitions; Keep uses 2,254 continuation calls and Drop 1,887, with 91 and 87 native successes. Its Keep arms supply the 132 confirmation recordings replayed in Section~\ref{sec:econ-mechanical-policies} and Appendix~\ref{app:econ-mechanical-policies}, and the trace in Figure~\ref{fig:econ-traces}. Its live Keep/Drop contrast is not used in the body.

\subsection{What the illustrated trace shows}
\label{app:econ-trace-sources}

\begin{figure*}[t]
\centering
\includegraphics[width=.94\textwidth]{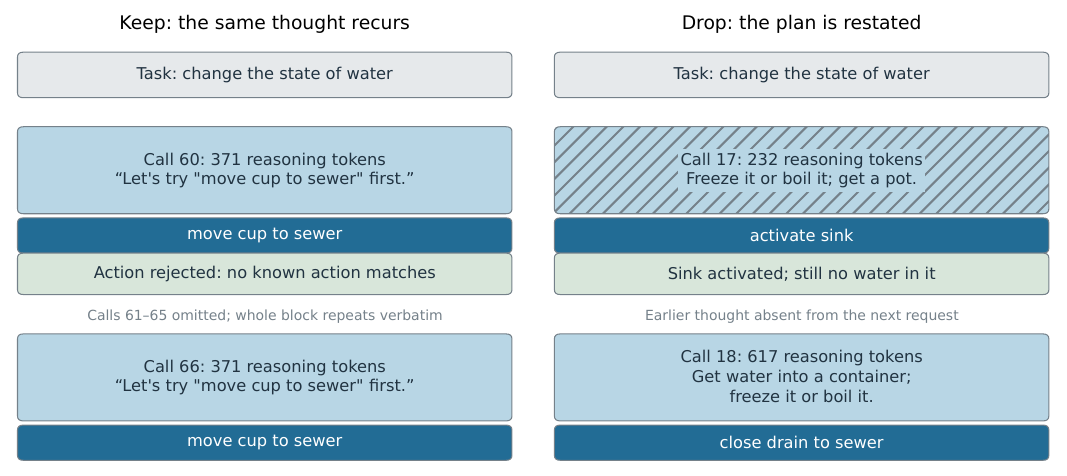}
\caption{\textbf{Retention changes future work.} One hosted Qwen3.7-Plus ScienceWorld pair: Keep repeats an identical 371-token block and action at calls 60 and 66; Drop restates a plan after earlier reasoning is removed (232 and 617 tokens). The Keep sentence is quoted; Drop text is paraphrased. The same retention intervention permits very different continuation patterns. Appendix~\ref{app:econ-trace-sources} identifies the public task and calls.}
\label{fig:econ-traces}
\end{figure*}

Figure~\ref{fig:econ-traces} shows ScienceWorld \texttt{change-\allowbreak the-\allowbreak state-\allowbreak of-\allowbreak matter-\allowbreak of}, variation 0, repetition 3. The displayed Keep excerpts are calls 60 and 66; Drop excerpts are calls 17 and 18.

The requests confirm that Drop's next call omits previous reasoning. Keep repeats an identical 371-token block, including ``Let's try \texttt{move cup to sewer} first.'' Drop's two blocks contain 232 and 617 reasoning tokens; their displayed wording is paraphrased. The excerpts illustrate the two continuation patterns shown in Figure~\ref{fig:econ-traces}.

\subsection{Provider retention rules}
The server's retention rule also varies. OpenAI documents current-turn reasoning as the default before GPT-5.6; GPT-5.6 defaults to all available compatible turns, with a request parameter selecting either mode \citep{openai2026reasoning}; Claude Sonnet 4.5 strips earlier-turn thinking, while Opus 4.5 and Sonnet 4.6 retain returned thinking by default and bill it as input \citep{anthropic2026extendedthinking}; Qwen3.6 retains thinking from the latest user turn by default and adds a switch to retain earlier turns \citep{qwen2026preservethinking}.

\section{Chained tasks and context compaction}
\label{sec:econ-client-factorial}

\subsection{Protocol and observed population}

All three completed cohorts use hosted \texttt{qwen3.7-plus-2026-05-26}, temperature 0.7, with native thinking enabled. The original cohort has four independent executions of the same ordered task stream: server-log analysis, a delayed-feedback bandit, grid pathfinding, a multi-labeller implementation, a SMILES data task, and keyless signing. A new task receives a fresh execution sandbox and its native verifier; the agent's history carries forward. This separates task-environment state from session context. Sequential independent tasks in one uninterrupted session are also used by \citet[Sec.~6.2]{arxiv260611213} to exercise context management.

The client policies are automatic Summary and the zero-shot CLM editing prompt \citep{shao2026contextlanguagemodels}. Summary activates at 24,000 tokens, 75\% of the 32,000-token input budget. CLM exposes the context as a text file editable through the single \texttt{bash} tool; a harness guard requires retaining the current task briefing. Keep returns native reasoning; Drop removes historical native reasoning and sets the provider's retention flag accordingly. The task order, model, budget and tool contract are shared. Each cell in the original cohort contains one six-task chain, with history shared across its tasks. All four chains submit all six tasks. Summary invokes four and two compactions under Keep and Drop; CLM applies two and three substantive edits.

The first expansion adds two fresh five-task streams, each run under all four conditions. The first contains Hydra/Slurm debugging, Maven dependencies, an authentication race, publisher analysis and competitive programming; the second contains a Merkle-tree CLI, sales statistics, pandas/NumPy analysis, an OkHttp crash and a housing-prediction API. Task selection uses a frozen seeded order after independent qualification and exclusions; model outcomes do not select replacements. Each task has an unchanged image across arms, a fresh container and verified history continuity. All eight chains submit all five tasks and compact before continuing.

The first expansion retains the model, temperature, Summary trigger and 32K working-context budget. Its output allowance increases from 16,384 to 65,536 tokens; command and effective task limits increase to 600 and 5,400 seconds, and a 131,072-token transport allowance accommodates compaction after a large output. A context-file byte-preservation repair precedes the fresh runs. The harness can roll back recent history on overflow: CLM Keep experiences 11 such resets, counted separately from eight substantive edits. We report the cohorts separately because these changes accompany the new task streams.

The second completed expansion adds two more five-task streams. One contains environment setup, IoT registration, log summarization, EKF localization and a to-do API; the other contains convolutional layers, ACL permissions, React/TypeScript debugging, supply-chain fulfillment and API-permission canonicalization. It keeps the 65,536-token output allowance and 32K working-context budget. Above that budget, the harness preserves history and blocks ordinary task actions until compaction restores room. All eight chains submit every task, compact and continue, with zero automatic history rollbacks or generation-cap endings. The usual 10,000-character tool-display bound clips 15 observations; full raw outputs are retained.

An earlier three-task, six-arm pilot did not activate compaction: the peak input was 20,451 tokens, below the Summary trigger. The reported cohorts use the later chained-task protocol. Base does not supply a complete six-task chain in this collection: Keep completes two tasks, both passing, and Drop completes four, three passing. Unattempted tasks are not scored as failures and these partial chains are not pooled with the four completed compaction cells.

\subsection{Usage, API cost, and native grading}

We independently sum 1,598 successful API responses (301 original, 544 first expansion and 753 second expansion), checking unique call identifiers, request retention flags, usage records and per-task native outcomes. The recorded tariff is \$0.40 per million uncached input tokens, \$0.08 per million cached input tokens, and \$1.60 per million output tokens. For prompt, cached and output counts $P,R,O$, the charge is
\[
 C_{\rm API}=10^{-6}\{0.40(P-R)+0.08R+1.60O\}.
\]
Reasoning is a subset of output and is counted once. The original four chains total \$1.12261288; the first expansion totals \$1.83512024 and the second \$2.80960880. The totals include uncached input, cached input and output at that tariff. Summary-production calls, unsuccessful actions and calls preceding failed native grades remain in the total.

\begin{table}[htbp]\centering\small
\begin{tabular}{@{}lrrrrrr@{}}\toprule
Condition & Passed & USD & Calls & Output tokens & Compactions & Rollbacks \\\midrule
\multicolumn{7}{l}{\textit{Original: one six-task stream}} \\
Summary Keep & 5/6 & 0.3168 & 80 & 77,609 & 4 & 0 \\
Summary Drop & 5/6 & 0.2386 & 77 & 58,080 & 2 & 0 \\
CLM Keep & 5/6 & 0.1959 & 55 & 32,649 & 2 & 0 \\
CLM Drop & 4/6 & 0.3713 & 89 & 81,119 & 3 & 0 \\
\midrule
\multicolumn{7}{l}{\textit{Expansion 1: two five-task streams}} \\
Summary Keep & 9/10 & 0.4130 & 142 & 91,996 & 4 & 0 \\
Summary Drop & 8/10 & 0.3677 & 139 & 70,017 & 3 & 0 \\
CLM Keep & 10/10 & 0.7389 & 157 & 114,241 & 8 & 11 \\
CLM Drop & 9/10 & 0.3155 & 106 & 55,430 & 2 & 0 \\
\midrule
\multicolumn{7}{l}{\textit{Expansion 2: two five-task streams}} \\
Summary Keep & 7/10 & 0.8108 & 207 & 187,190 & 11 & 0 \\
Summary Drop & 9/10 & 0.5600 & 158 & 122,699 & 6 & 0 \\
CLM Keep & 9/10 & 0.6917 & 203 & 108,483 & 10 & 0 \\
CLM Drop & 9/10 & 0.7470 & 185 & 104,000 & 5 & 0 \\
\bottomrule\end{tabular}
\caption{Completed hosted cohorts, reported separately. Output includes native reasoning once. Costs include summary calls, edits and failed task attempts. Compactions count automatic summaries or applied substantive edits; rollbacks are automatic harness history resets.}\label{tab:econ-chain-totals}
\end{table}

In the original cohort, CLM/Summary is $0.6182$ under Keep and $1.5559$ under Drop. The first expansion reverses that ordering: $1.7891$ and $0.8581$, respectively. Its Drop/Keep ratios are $0.8903$ for Summary and $0.4270$ for CLM. Both clients pass one fewer task under Drop; CLM passes one more than Summary under each retention setting. In the second expansion, CLM/Summary is $0.8531$ under Keep and $1.3339$ under Drop. Dropping reasoning lowers Summary's expenditure by 30.9\% and raises CLM's by 8.0\%. CLM passes nine tasks under each retention rule; Summary passes seven under Keep and nine under Drop. Sequential tasks share history, so the units are the five matched task streams, with cohorts shown separately.

The second expansion also includes a manual instruction audit alongside native grading. Six native passes have additional findings: three ACL executions create prohibited temporary files; CLM Drop changes supplied React tests and misses a global naming rule in API canonicalization; Summary Keep implements a narrower inheritance rule than specified. An already-failing CLM Keep React execution also changes test setup. These seven findings remain distinct from the frozen native grades; all associated costs and tasks remain in the comparison.

Later interrupted attempts remain outside these completed-cohort comparisons. One ten-task batch completes six of eight chains before its fixed wall-time limit; its registry task also lacks a promised input, and the native certificate instances differ across arms. Subsequent attempts stop on a provider rejection or harness errors. Their submitted tasks, failed grades and costs remain archived without combining partial chains into a completed replication. The totals above use the original cohort and both completed expansions.

\subsection{Conditional mechanical--continuation replay}
\label{app:econ-chain-replay}

The replay in Table~\ref{tab:econ-chain-replay} uses the original six-task cohort. Let $K,D$ denote Keep and Drop. The reference is the Keep recording: $\Delta M=C(K;\xi_K)-C(D;\xi_K)$, $\Delta B=C(D;\xi_K)-C(D;\xi_D)$. All terms use the same declared renderer and cache rule. The API-dollar totals above are not inserted into this counterfactual decomposition.

The replay tokenizes canonical role delimiters, thinking blocks and tool-call serialization with the archived Qwen3 tokenizer. It holds generated replacements, compaction timing, tool inputs, task order and outputs fixed. Counterfactual Drop removes historical native reasoning fields, including those sent to summarization. For migrated CLM text, it removes reasoning only when an entire intact rendered turn matches a previous response from that chain. Authored or partly rewritten replacements remain fixed. These rendering and erasure rules define the counterfactual used in the decomposition.

Reuse is the longest exact-token prefix rounded down to 16-token blocks. Prompt-only reuse retains the preceding prompt; generated-state reuse retains the canonical prompt and completion, capped at the preceding charged $P+G$. Generation uses provider output counts, with an emitted-minus-one sensitivity. The reference prices use $P^*=123{,}850$ and $w=1$; they project the recorded work onto the hybrid reference architecture, rather than identify the hosted model's coefficients.

\begin{table}[htbp]\centering\small
\begin{tabular}{@{}llrrr@{}}\toprule
Client & $\rho$ & $\Delta M$ & $-\Delta B$ & $\Delta C$ \\\midrule
Summary & 0 & +21.73 & -19.47 & +41.19 \\
Summary & 1 & +3.60 & -23.98 & +27.58 \\
CLM & 0 & +11.95 & +62.99 & -51.04 \\
CLM & 1 & -8.82 & +77.76 & -86.58 \\
\bottomrule\end{tabular}
\caption{Declared compaction replay at $P^*=123{,}850$, $w=1$. Entries are percentages of replayed Keep cost, not API expenditure. Policy 1 is Keep and the baseline Drop, so positive $\Delta C$ means Keep costs more; positive $-\Delta B$ is a continuation saving, and $\Delta C=\Delta M-(-\Delta B)$. Rounding accounts for differences in displayed sums.}\label{tab:econ-chain-replay}
\end{table}

Mean per-call prompt discrepancies against provider receipts are $+0.56\%$, $+4.56\%$, $+0.48\%$, and $+5.89\%$ for Summary Keep/Drop and CLM Keep/Drop. Replay checks confirm that erasure leaves archived Drop requests unchanged, generation stays fixed in mechanical replay, all reusable prefixes fit, and $\Delta M+\Delta B=\Delta C$. Across the saved sensitivities in exchange rate, generation price, cache branches, final-token convention and erasure scope, Summary Drop/Keep lies in $[0.569,0.739]$ and CLM in $[1.510,2.286]$. These are reconstruction ranges, not confidence intervals. CLM's mechanical component changes sign across them, so its value always carries a serving label.

In the original cohort, the CLM cost gap between Drop and Keep is not a compaction effect: Drop makes 34 more model calls than Keep over the stream, only one of them a context-management call, and about 98\% of its extra output tokens fall outside context management. Compaction counts follow the applied context state, not shell messages, since several attempted edits were rejected by the briefing guard.

\subsection{Visible thoughts and native reasoning}
\label{app:econ-thought}

Visible deliberation and native reasoning are separate channels. ReAct interleaves explicit reasoning traces and actions in the model's ordinary output \citep{yao2023react}. The released CLM zero-shot prompt likewise requests a visible \texttt{THOUGHT} before an action \citep{shao2026contextlanguagemodels}. A provider's native reasoning field is a different object with its own rendering and retention rules. A request can therefore preserve visible thought while omitting the native reasoning field.

Our compaction harness preserves visible assistant text under both retention settings, except where a client edit or summary replaces it. In the original CLM chains, 53 of 55 Keep responses and 25 of 89 Drop responses contain a visible \texttt{THOUGHT}; it appears in historical context in 54 and 62 requests, respectively. Native reasoning appears in historical fields in 54 Keep requests and zero Drop requests. The original cost reversal thus occurs while visible thoughts remain available to the Drop agent. In the first expansion, historical visible \texttt{THOUGHT} occurs in 115 of 157 CLM Keep requests and 76 of 106 Drop requests; no Drop request contains a historical native reasoning field. The first expansion favors Drop on cost despite the same distinction between channels. In the second expansion, visible \texttt{THOUGHT} appears in 71 of 203 CLM Keep requests and 149 of 185 CLM Drop requests; historical native reasoning is again absent from all Drop requests. Visible-thought content and frequency vary with the continuation, so this comparison evaluates the complete retention policies.

\paragraph{An actual request example.}
In the first server-log task, the first historical assistant message in each CLM arm contains exactly the same visible text:
\begin{quote}\small
\texttt{THOUGHT:} Let me start by examining the questions file and a sample of the access log to understand what I'm working with.
\end{quote}
In the Keep request, a separate \texttt{reasoning\_content} field says: ``Let me start by examining the input files to understand what I'm working with.'' In the Drop request that field is absent. Both excerpts are copied from the actual request payloads. Their shared visible sentence makes the distinction concrete: the same visible thought can travel with or without native reasoning. Later context edits change the respective histories.

Across all three cohorts, CLM retention arms receive context-size feedback and generic editing guidance, including that there is no penalty for extra turns. Neither receives the cost coefficients, cache receipts, or a computed rebuilding charge. The experiment evaluates transfer of this zero-shot prompt across retention settings.

\section{Fresh sessions and carried history}
\label{app:econ-sequential}

\paragraph{Design.}
Qwen3-8B runs in BF16 through vLLM 0.8.5 on an A6000. Each of 32 matched chains contains four distinct task families and then repeats its first task and variation. Tasks are sampled from the 14 ScienceWorld families solved at least once in the earlier 8B fork collection. Variations come from the benchmark's test split. The task and arm-order seed is zero. Each chain executes all five arms in a shuffled order. Before each task, the environment loads its task and variation and calls reset; the worker environment is reused. The budget is ten model calls per task, 3,000 generated tokens per call, and a 32,768-token context, with a 3,072-token generation reserve. The run uses temperature zero and ten concurrent workers.

\paragraph{Policies.}
Fresh starts a new transcript at every task. Carry retains completed tasks, appending the previous task's final observation and the new task briefing. Keep renders earlier reasoning; Drop removes its text and leaves empty thinking delimiters. The Clear option uses Keep and adds a system instruction offering the action \texttt{clear memory}, which removes all earlier tasks while retaining the current one. This arm changes both the instruction and the available action. If the input exceeds the context budget less its reserve, the oldest task is removed. This overflow rule activates once in Carry/Keep and four times in the clearing arm; it never activates in the other arms.

\begin{table}[ht]
\centering\small
\begin{tabular}{lrrrrr}
\toprule
Policy & Calls & Solved & Cost / fresh & New tasks & Clear actions \\
\midrule
Fresh / Keep & 1,094 & 47/160 & 1.00 [1.00, 1.00] & 30/96 & 0 \\
Carry / Keep & 1,139 & 34/160 & 0.60 [0.41, 0.87] & 17/96 & 0 \\
Clear option / Keep & 1,207 & 42/160 & 0.98 [0.53, 1.67] & 25/96 & 45 \\
Fresh / Drop & 1,199 & 33/160 & 1.00 [1.00, 1.00] & 26/96 & 0 \\
Carry / Drop & 1,222 & 19/160 & 0.43 [0.36, 0.52] & 13/96 & 0 \\
\bottomrule
\end{tabular}
\caption{All five policies on the same 32 chains. Cost is the ratio of summed generated-state-reuse replay cost to Fresh under the same reasoning-retention rule. Intervals resample whole matched chains. ``New tasks'' counts positions two through four; the first and repeated task are excluded from that column.}
\label{tab:econ-chained-history}
\end{table}

\paragraph{Accounting and uncertainty.}
All 5,861 model calls reconcile with the task summaries: prompt and generation counts, recorded arithmetic, terminal scores and success flags. Costs in Figure~\ref{fig:econ-chained-history} use the archived tokenizer replay with $c=13.9$ GFLOPs/token, $\beta=589{,}824$ FLOPs/pair and $w=1$. The renderer reconstructs the task transcript and canonical parsed completion. Reuse is the longest common prefix with the preceding prompt and completion, rounded to 16-token blocks and leaving the final prompt token to evaluate. Prompt lengths match the recorded lengths in all calls. The clearing arm's input boundary is reconstructed by matching the recorded prompt length. These are specified cache reconstructions. The separate server cache receipt does not define this comparison.

Pricing the original logged counts instead gives Carry/Fresh ratios of 0.60 under Keep and 0.45 under Drop, and Clear option/Fresh of 0.99; canonical replay gives 0.60, 0.43 and 0.98. Each ratio divides total cost over the complete panel. For the clearing arm, the median individual-chain ratio is 0.67, while the ratio of totals is 0.98. The main figure uses the latter consistently. Percentile 95\% intervals use 4,000 bootstrap draws with seed 20261013, retaining all task positions and arms within each sampled chain.

\paragraph{Task position and generated work.}
On the three subsequent new tasks, Carry solves 17/96 under Keep against Fresh's 30/96, and 13/96 under Drop against 26/96. Both differences are $-13.5$ percentage points, with chain-bootstrap intervals $[-22.9,-4.2]$ and $[-21.9,-5.2]$. On the repeated task, the Keep counts are 9/32 versus 8/32 and the Drop counts 3/32 versus 4/32. The clearing option solves 25/96 new tasks and invokes its action 45 times, 42 at the first call of a new task. Sixteen chains never clear. At the 128 transitions to a later task, the first call has empty reasoning 84 times under Carry/Keep and 117 under Carry/Drop, against zero in either Fresh arm. These counts describe the continuations produced under each complete history policy.

\section{Mechanical policy replay across price regimes}
\label{app:econ-mechanical-policies}

\paragraph{Recorded continuation and policies.}
The replay uses the Keep arms of the hosted ScienceWorld collection in Appendix~\ref{app:econ-hosted}. Its original development partition contains 27 traces from five task families; the confirmation partition contains 132 traces from the remaining 25 families, with 1,900 calls. All 159 traces and 2,254 calls enter the reconstruction checks. The body reports the confirmation partition. The original partition precedes the window-replay analysis; the newer periodic-deletion schedule is evaluated on the same saved partitions.

Each reasoning block, answer and subsequent input increment receives a distinct segment identity. Reasoning and answer lengths come from usage receipts. The input increment is the next prompt length minus the preceding prompt and emitted output lengths. These increments include external inputs and message framing. Accumulating the segments reproduces every archived Keep prompt length. The initial shared prompt remains unchanged. Keep all retains all subsequent reasoning; Drop immediately removes each block before its next use; the sliding windows retain the latest one or five blocks; periodic deletion removes all completed reasoning before calls 6, 11, and so on. All answers and external inputs remain. Replacement text and replacement-generation calls are absent from these deletion schedules.

\paragraph{Cache and price.}
At each call, reuse is the longest contiguous prefix matching any prior cached segment sequence for that policy. Generated-state reuse stores the prompt and evaluated output positions; prompt-only reuse stores the prompt. The final emitted token has no cached state, so an output of $O_t$ tokens uses $G_t=O_t-1$ decode positions. The first call reuses its recorded initial prefix in both cache modes. No-reuse sets every $R_t=0$. Cache prefixes are rounded down to blocks of 16 tokens in the figures, with one- and 128-token sensitivities calculated separately. Cache capacity is unrestricted and entries are never shared between policies.

The reference coefficients are $c=2{,}818{,}572{,}288$ and $\beta=229{,}376$ from the Qwen3-1.7B projection account, hence $P^*=12{,}288$. They price the recorded work on a specified architecture. The sweep varies $P^*$ from 1K to 2M at 121 logarithmically spaced values, with $w=1,2,3,4,5$. Ratios of panel totals remove the common absolute coefficient $\beta$.

\begin{table}[ht]
\centering
\small
\begin{tabular}{lrrrr}
\toprule
Reasoning policy & Prompt volume & Full reuse & Prompt-only & No reuse \\
\midrule
Keep all & 1.000 & 1.000 & 1.000 & 1.000 \\
Drop immediately & 0.976 & 0.983 & 0.955 & 0.946 \\
Keep latest 1 & 0.978 & 1.861 & 1.809 & 0.949 \\
Keep latest 5 & 0.984 & 4.542 & 4.416 & 0.960 \\
Clear every 5 & 0.978 & 1.598 & 1.560 & 0.948 \\
\bottomrule
\end{tabular}
\caption{Policy totals relative to Keep all on the 132 confirmation continuations. Prompt volume counts all submitted tokens. The three cost columns use the reference prices, $w=1$ and 16-token cache blocks. Across $P^*\in[1\mathrm{K},2\mathrm{M}]$ and $w\in[1,5]$ with generated-state reuse, keeping the latest five blocks costs 3.97 to 4.56 times Keep and immediate deletion stays within 2\% of Keep.}\label{tab:econ-mechanical-policies}
\end{table}

Figure~\ref{fig:econ-mechanical-policies} displays the schedule ratios at the reference exchange rates of Qwen3-30B-A3B, Qwen3-8B and Qwen3.6-27B, with $w=1$.

\paragraph{What changes with the decoding premium.}
All schedules share $G$. For policies $i$ and $j$, Equation~\ref{eq:econ-cost} therefore gives $C_i-C_j=c(N_i-N_j)+\beta(Q_i-Q_j)$, independent of $w$. Raising $w$ adds the same expenditure to both policies, moving their cost ratio toward one. Changing $P^*$ or the reusable prefix changes the mechanical comparison itself. This isolates edit placement and timing from differences in generated work.

\paragraph{Pricing additional generation.}
The right panel of Figure~\ref{fig:econ-mechanical-policies} prices one evaluated output position with $H=32{,}000$ attended keys, returned unchanged at the same position in the next prompt. Exact token-granular caching preserves its state under $\rho=1$; prompt-only reuse reprocesses it once. Its own write-and-return cost, divided by the cost of one input read at that position, is
\begin{align}
 &\frac{wc+\beta H+(1-\rho)(c+\beta H)}{c+\beta H}\nonumber\\
 &\qquad=\frac{(w+1-\rho)P^*+(2-\rho)H}{P^*+H}.
\end{align}
As $P^*/H$ grows, the ratio tends to $w+1-\rho$: one or two input reads at $w=1$, five or six at $w=5$. Later queries attending to this token contribute carry separately. This calculation uses the stated two evaluations; the left panels use the recorded continuations. The generation term shared by those recordings cancels in a schedule difference, while additional generation incurs the price shown on the right.

\paragraph{Checks and sensitivity.}
The segment-prefix implementation agrees with explicit token-identity sequences in 1,800 synthetic comparisons; attention counts also agree with position-by-position sums. Recomputed counts match all 1,908 archived trace--policy--alignment combinations for the four original policies. At the reference prices, changing alignment from one to 128 tokens leaves the five-block window between 4.53 and 4.54 times Keep in the confirmation partition. On the development partition it costs approximately 4.15 times Keep. Immediate deletion and Keep remain within two percent in both partitions under generated-state reuse. These results price the stated deletion schedules on our ScienceWorld continuations; the SWE-AGILE comparison supplies the sliding-window design, including its distinct use of digests in the published method.
\end{document}